\documentclass[a4paper,fleqn]{article}
\usepackage{amsmath}
\usepackage{arxiv}
\usepackage[numbers]{natbib}
\usepackage{bm} 
\usepackage{amssymb}
\usepackage{subcaption}
\usepackage{todonotes}
\usepackage{setspace}
\usepackage{float}
\usepackage{hyperref}       
\usepackage{url}            
\usepackage{ulem}
\usepackage{doi}
\newcommand{\rw}[1]{\textcolor{blue}{#1}}
\newcommand{\del}[1]{\textcolor{red}{\sout{#1}}}
\renewcommand{\del}[1]{}
\renewcommand{\rw}[1]{#1}

\title{The Deep Arbitrary Polynomial Chaos Neural Network or how Deep Artificial Neural Networks could benefit from Data-Driven Homogeneous Chaos Theory}

\author{ \href{https://orcid.org/0000-0003-4676-5685}{\includegraphics[scale=0.06]{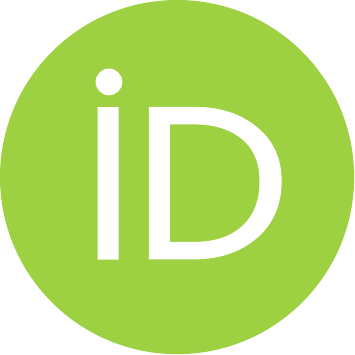}
	\hspace{1mm}Sergey Oladyshkin}\\
	Department of Stochastic Simulation and Safety Research for Hydrosystems,\\
	Institute for Modelling Hydraulic and Environmental Systems, Stuttgart Center for Simulation Science,\\
	University of Stuttgart,
	Pfaffenwaldring 5a, 70569 Stuttgart, Germany\\
	\texttt{Sergey.Oladyshkin@iws.uni-stuttgart.de} \\
	\And 
	\hspace{1mm}Timothy Praditia\\
		Department of Stochastic Simulation and Safety Research for Hydrosystems,\\
	Institute for Modelling Hydraulic and Environmental Systems, Stuttgart Center for Simulation Science,\\
	University of Stuttgart,
	Pfaffenwaldring 5a, 70569 Stuttgart, Germany\\
	\And 
	\href{https://orcid.org/0000-0003-0360-5307}{\includegraphics[scale=0.06]{orcid.pdf}\hspace{1mm}Ilja Kröker} \\
	Department of Stochastic Simulation and Safety Research for Hydrosystems,\\
	Institute for Modelling Hydraulic and Environmental Systems,
	Stuttgart Center for Simulation Science,\\
	University of Stuttgart,
	Pfaffenwaldring 5a, 70569 Stuttgart, Germany\\
	\And
	\hspace{1mm}Farid Mohammadi\\
	Department of Hydromechanics and Modelling of Hydrosystems,\\ Institute for Modelling Hydraulic and Environmental Systems,\\ University of Stuttgart,
	Pfaffenwaldring 61, 70569 Stuttgart, Germany\\
	\And
	\href{https://orcid.org/0000-0003-2583-8865}{\includegraphics[scale=0.06]{orcid.pdf}\hspace{1mm}
	Wolfgang Nowak}\\
		Department of Stochastic Simulation and Safety Research for Hydrosystems,\\
	Institute for Modelling Hydraulic and Environmental Systems,
	Stuttgart Center for Simulation Science,\\
	University of Stuttgart,
	Pfaffenwaldring 5a, 70569 Stuttgart, Germany\\
	\And
	\hspace{1mm}Sebastian Otte\\
	Neuro-Cognitive Modeling, Computer Science Department,\\ University of T\"ubingen,
	Sand 14, 72076 T\"ubingen, Germany
	}

\renewcommand{\shorttitle}{Deep Arbitrary Polynomial Chaos Artificial Neural Network}

\begin{document}

\let\WriteBookmarks\relax
\def\floatpagepagefraction{1}
\def\textpagefraction{.001}



%


\maketitle


\begin{abstract}
Artificial Intelligence and Machine learning have been widely used in various fields of mathematical computing, physical modeling, computational science, communication science, and stochastic analysis. Approaches based on Deep Artificial Neural Networks (DANN) are very popular in our days. Depending on the learning task, the exact form of DANNs is determined via their multi-layer architecture, activation functions and the so-called loss function. However, for a majority of deep learning approaches based on DANNs, the kernel structure of neural signal processing remains the same, where the node response is encoded as a linear superposition of \del{neurons} \rw{neural activity}, while the non-linearity is triggered by the activation functions. In the current paper, we suggest to analyze the neural signal processing in DANNs from the point of view of  homogeneous chaos theory as known from polynomial chaos expansion (PCE). From the PCE perspective, the (linear) response on each node of a DANN could be seen as a $1^{st}$ degree multi-variate polynomial of single neurons from the previous layer, i.e. linear weighted sum of monomials. From this point of  view, the conventional DANN structure relies implicitly (but erroneously) on a Gaussian distribution of neural signals. Additionally, this view revels that \rw{by design} DANNs do not necessarily fulfill any orthogonality or orthonormality condition for a majority of data-driven applications. Therefore, the prevailing handling of neural signals in DANNs could lead to redundant representation as any neural signal could contain some partial information from other neural signals. To tackle that challenge, we suggest to employ the data-driven generalization of PCE theory known as arbitrary polynomial chaos (aPC) to construct a corresponding multi-variate orthonormal representations on each node of a DANN. Doing so, we generalize the conventional structure of DANNs to Deep arbitrary polynomial chaos neural networks (DaPC NN). They decompose the neural signals that travel through the multi-layer structure by an adaptive construction of data-driven multi-variate orthonormal bases for each layer.  Moreover, the introduced DaPC NN provides an opportunity to go beyond the linear weighted superposition of single neurons on each node. Inheriting fundamentals of PCE theory, the DaPC NN offers an additional possibility to account for high-order neural effects reflecting simultaneous interaction in multi-layer networks. Introducing the high-order weighted superposition on each node of the network mitigates the necessity to introduce non-linearity via activation functions and, hence, reduces the room for potential subjectivity in the modeling procedure. \rw{Although the current DaPC NN framework has no theoretical restrictions on the use of activation functions}. The current paper also summarizes relevant properties of DaPC NNs inherited from aPC as analytical expressions for statistical quantities and sensitivity indexes on each node. We also offer an analytical form of partial derivatives that could be used in various training algorithms. Technically, DaPC NNs require similar training procedures as conventional DANNs, and all trained weights determine automatically the corresponding multi-variate data-driven orthonormal bases for all layers of DaPC NN. The paper makes use of three test cases to illustrate the performance of DaPC NN, comparing it with the performance of the conventional DANN and also with plain aPC expansion. Evidence of convergence over the training data size against validation data sets demonstrates that the DaPC NN outperforms the conventional DANN systematically. Overall, the suggested re-formulation of the kernel network structure in terms of homogeneous chaos theory is not limited to any particular architecture or any particular definition of the loss function. The DaPC NN Matlab Toolbox is available online and users are invited to adopt it for own needs. 
\end{abstract}










\section{Introduction}
\label{intro}

During the last decades, Artificial Intelligence (AI) and Machine learning (ML) have been widely used in various fields of mathematical computing, physical modeling, computer science, geosciences, communication science, and stochastic analysis. The terminology AI has been suggested by John McCarthy in 1956 as a neutral title of a Dartmouth workshop \cite{mccarthy1988review} to distinguish the research field from cybernetics and also to escape the influence of its originator Norbert Wiener \cite{wiener1948cybernetics}. The closely related term ML has been introduced later in 1959 by Arthur Samuel, where the author explored the logical rules of the game of checkers \cite{samuel1959some}.  Originally, AI and ML have been focused on learning strategies employing logical rules, which were often formalized using an apparatus of discrete mathematics and graph theory. However, with increasing computational power \cite{miikkulainen2019evolving} and data availability \cite{najafabadi2015deep}, the fields of AI and ML today employ a much broader spectrum of approaches that originate from stochastic analysis, cybernetics, geosciences, information theory and other disciplines.  

In particular, approaches based on Deep Artificial Neural Networks (DANN) introduced in cybernetics by Alexey Ivakhnenko and Valentin Lapa in 1967 \cite{ivakhnenko1967cybernetics} are currently very popular in AI and ML \rw{(see e.g. \cite{schmidhuber2022annotated} for a detailed historical recapitulation)}. DANNs generalize the concept of Artificial Neural Networks (ANN) suggested by Warren McCulloch and Walter Pitts in 1943 \cite{mcculloch1943logical} to multi-layer structures \rw{, i.e. deep ANN}. This form of deep learning also gained a strong visibility in society, providing solutions for \rw{a} broad variety of tasks including recognition of images \cite{tian2020image}, videos \cite{ciaparrone2020deep},  voice \cite{arik2017deep} and text \cite{shi2016end}. 

Depending on the modelling task, the exact form of the so-called loss function \cite{Goodfellow2016deeplearning} is usually specified to determine the final DANN representation. Various loss functions can be found in literature, suitable for classification tasks \cite{ballard1982computervision,krizhevsky2017cnn}, Bayesian interpolation \cite{mackay1992bayesian} or physical regularization \cite{jia2021pgnn,praditia2020thermo}.  In addition, multiple DANN layers and corresponding neural connections can be  customized in various ways, turning DANN into convolutional \cite{buda2018systematic,rawat2017deep}, recurrent \cite{hochreiter1997long,xiao2018nonlinear} or other desired architectures \cite{karim2019multivariate}. Due to the magnificent number of works based on the DANN representation, the corresponding literature can hardly be covered in a research paper, and the authors refer to literature \rw{\cite{hassoun1995fundamentals,graupe2013principles}} and reviews \rw{\cite{bouwmans2019deep,schmidhuber2015deep}} for further information.

The keystone of DANNs can be seen as a certain type of non-linear function that maps from input in $\mathbb{R}^n$ to output in $\mathbb{R}$. To construct such a function, DANNs use a multi-layered approach, where each node of a layer is a linear combination of non-linear \rw{univariate} functions, known as activation functions. Overall, this yields a chain-rule interaction of neurons in multi-layered architecture \cite{anthony2009neural}. However, there is an alternative branch of approaches that also maps $\mathbb{R}^n$ to $\mathbb{R}$ using linear combinations of non-linear kernel functions or vectors and, hence, could be seen as one-layer approaches. This alternative ML branch consists of polynomial chaos expansions (PCE) introduced in 1938 \cite{Wiener1938}, Kriging introduced in 1951 \cite{krige1951statistical} that is also known as Gauss\rw{ian} process emulator/regression (GPE: \cite{williams2006gaussian}) or Wiener–Kolmogorov prediction \cite{cressie1993spatial}, Support vector regression (SVR) introduced in 1974 \cite{vapnik1974theory} and Relevance vector machines (RVM) introduced in 2000 \cite{tipping2000relevance}. The common fundamental between PCE, GPE, SVR and RVM can be found in \cite{KaiSurrogate}. 

Regardless of the different mathematical definitions and structures, both one- and multi-layered ML approaches construct the \del{finil} \rw{final} non-linear representation by determining all their unknown constants. These constants are known as coefficients (terminology for PCE, GPE, SVR and RVN) or weights (terminology for ANN and DANN). For one-layer approaches, the response is established by solving a linear system of equations that encodes the linear combination of non-linear basis functions (polynomials, kernels, vectors, etc.). For multi-layer approaches, the response is constructed by solving a non-linear system of equations that reflects the so-called neural signal processing. 

\begin{figure}[!t]
	\centering
     \begin{subfigure}{0.45\textwidth}
         \centering
         \includegraphics[width=\textwidth]{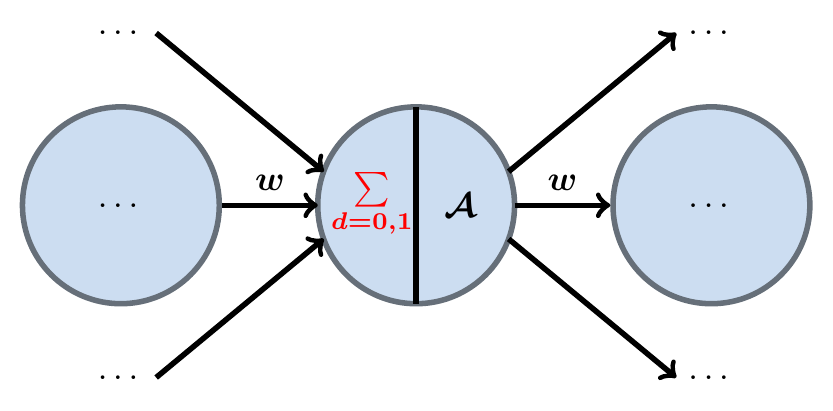}
         \caption{}
         \label{fig:dann}
     \end{subfigure}
     \hfill
     \begin{subfigure}{0.45\textwidth}
         \centering
         \includegraphics[width=\textwidth]{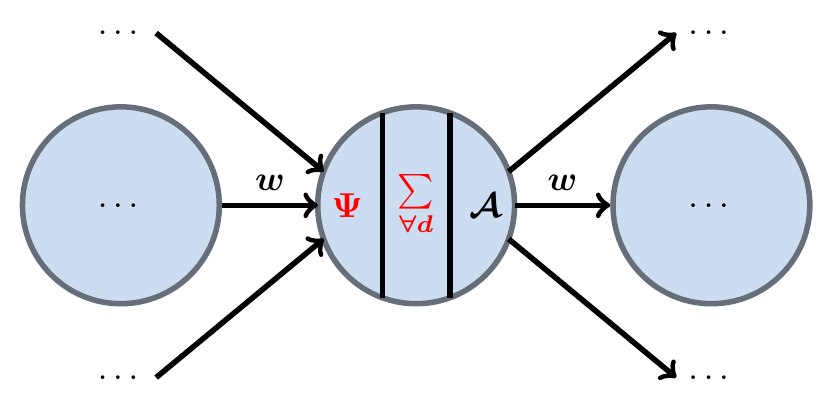}
         \caption{}
         \label{fig:dapc}
     \end{subfigure}
        \caption{Sketches of the neural processing in (a) DANN and (b) DaPC NN}
        \label{fig:dann_dapc}
\end{figure}

The current paper does not aim at discussing the pros and contras behind various approaches to map $\mathbb{R}^n$ to $\mathbb{R}$. We will rather pay attention to how one-layer findings could be helpful for the multi-layered structure of DANNs. Indeed, let us have a close look into the kernel structure of conventional DANNs. It considers the processing of the neural signal in one node as displayed in Figure \ref{fig:dann}. The response of each node (i.e. arrows leaving the central circle)  \del{is} contains linear weighted superposition of single neuron responses (i.e. the arrows entering the central circle) coming from the previous layer. In the figure, the weighting is represented by $w$ and the superposition by $\Sigma$. To obtain the final neuronal response, the superposition is passed through an activation function $\mathcal{A}$ that is usually non-linear. The corresponding weights $w$ of the linear superposition for the input to the featured neurons \del{is} \rw{are} to be found by training. However, such a linear weighted superposition of incoming neuron outputs on each node could lead to a redundant representation, as it is not necessarily satisfying orthogonality in signal processing. This aspect has been addressed in the literature on Support Vector Networks in 1995 \cite{cortes1995support} employing the SVR concept \cite{vapnik1974theory}. \rw{Also imposing orthogonality within the DANNs has been explored in the context of Recurrent Neural Networks \cite{mhammedi2017efficient} assuring the efficiency of the training procedure. The paper \cite{vorontsov2017orthogonality} highlights the benefit of using orthogonal weight matrices in Recurrent Neural Networks, as they preserve gradient norm during backpropagation making them highly desirable for optimization purposes. However, the imposition of hard constraints on orthogonality within Recurrent Neural Networks may negatively affect convergence speed \cite{vorontsov2017orthogonality} as applying Gaussian prior regularization may not be appropriate for many applications. Additionally, in order to overcome the difficulty of training Recurrent Neural Networks caused by vanishing and exploding gradients, a new approach have been addressed  that learns a unitary weight matrix with eigenvalues \cite{wisdom2016full,arjovsky2016unitary}, enabling optimization in the complex domain.} Nevertheless, accounting for orthogonality seems to be very promising, as it could mitigate redundancy in DANN representation \cite{wang2020orthogonal} and potentially could provide a better ability for generalization \cite{jia2019orthogonal}. 

Additionally, the actual non-linearity of DANNs is triggered by the non-linearity of the choosen activation functions. However, the choice of the activation functions is an extremely non-trivial task itself \cite{mhaskar1994choose} and can be very subjective \cite{sharma2017activation}, posing additional challenges for DANN users. Very recently, the work \cite{chrysos2020p} suggested to replace non-linear activation functions via non-linear polynomial representations, introducing co-called $\Pi$-Nets. The introduced $\Pi$-Nets consider high-order polynomial terms, which consistently improves the performance in discriminative and generative tasks for images and audio processing \cite{chrysos2020p}. Nevertheless, similar to conventional DANNs, $\Pi$-Nets do not yet consider orthogonality in processing the neural signal and, hence, could also lead to redundancy in representation. We argue that employing orthogonal (or even better orthonormal) decompositions for processing neural signals could be extremely relevant, especially for data-poor applications.

In the current work, we will take the reader on a journey to the early stages of ML. We will pay special attention to the PCE, introduced via the homogeneous chaos theory by Norbert Wiener in 1938 \cite{Wiener1938} as already mentioned above. The so-called non-intrusive version of the PCE \cite{Ghanem_Spanos_1991_SFEM_book,Lin2009} and its advanced extensions towards sparse quadrature \cite{Keese2003}, sparse \del{integration rules} \rw{regression} \cite{ahlfeld2016samba,blatman2008sparse} or multi-element approaches \cite{alkhateeb2017data,knr2012} gained popularity for surrogate building in computationally demanding modelling tasks \cite{foo_pcm_JCP2010,Olad_al_CG2009}. They all employ the idea of multi-variate polynomial representation. \rw{The fundamental concept of PCE theory lies }  \del{The kernel idea of PCE theory consists} in projection of functions onto a space spanned by orthonormal polynomials that capture the non-linear dependencies. 

Using the DANN vocabulary, the PCE could be seen as a one-layer ML approach, where the response is approximated as a linear combination of non-linear \rw{multivariate} orthonormal polynomial basis functions. At the same time, employing the PCE vocabulary, the response of one node in a DANN could be seen as a $1^{st}$ degree multi-variate polynomial of single neurons from the previous layer, i.e. linear weighted superposition, then passed through an activation function. To be more specific, the $1^{st}$ degree multi-variate polynomial representation in one node of a DANN consists of $1^{st}$ degree monomials that reflect the incoming neural signals and the \del{$0^{st}$} \rw{$0^{th}$} degree term that is known as bias. Obviously, this representation via linear weighted superposition of monomials does not necessarily fulfill the orthogonality or orthonormality condition that is targeted by PCE approach for the majority of applications. 

Such a non-orthogonal linear weighted superposition of incoming neuron outputs is a possibly redundant representation of the overall signal flow: any neural signal could contain some partial information from other neural signals. Similarly, the non-linear polynomial representation in $\Pi$-Nets \cite{chrysos2020p} could also profit from orthonormal decomposition of neural signals. Therefore, employing multi-variate orthonormal polynomial bases to process the neural signal on each node, as in the PCE approach, could be beneficial for both conventional DANN structures and for advanced $\Pi$-Nets. 
\rw{Another approach that combines variational autoencoder and PCE methods is expected to appear soon in the literature forming the PCE-Net  model \cite{shustin2022PCENet}.}

Unfortunately, due to the data-driven nature of all these networks, a direct transfer of classical PCE theory or its generalized extension \cite{xiu2002wiener} for constructing orthonormal representations on each node is not possible. The reason is that PCE requires knowledge of probability density functions for all inputs, and this knowledge is not available in the context of DANNs, where the probability density function for all inputs to all nodes in all layers would have to be known. 

Therefore, the current paper suggests to account for the data-driven nature of neural signals in DANNs and thus uses the data-driven generalization of PCE known as arbitrary polynomial chaos (aPC: \cite{OladNowak_RESS2012}) to construct corresponding multi-variate orthonormal representations on each node of DANNs.  Doing so, we generalize the conventional structure of DANNs to Deep arbitrary polynomial chaos neural networks (DaPC NN)\footnote{The authors of the current paper came across the interesting work under revision \cite{zheng2021minidatadriven} denoted as Deep aPCE that seems to be developed in parallel with the current DaPC NN work. In order to avoid any confusion for the reader, we would like to clarify that both works employ the data-driven fundamentals of aPC \cite{OladNowak_RESS2012}. However, regardless the similarity in the abbreviations, the Deep aPCE makes use of DANNs to construct the aPC expansion coefficients, but the DaPC NN employs aPC representations to construct a re-formulated DANN. We refer the reader to the original paper \cite{zheng2021minidatadriven} for more details.}. They decompose the neural signals traveling through the multi-layer structure through an adaptive construction of data-driven multi-variate orthonormal bases for each layer. Doing so, we provide an opportunity to go beyond the linear weighted superposition of single (\rw{univariate}) neurons on each node that is traditionally employed in various DANN architectures. 

Inheriting fundamentals of the PCE, the DaPC NN offers an additional possibility to account for high-order neural effects that reflect simultaneous interaction of neurons in the multi-layer network. Thus, the modeler is prompted to specify the DaPCE NN architecture through not only the number of layers, the number of nodes per layer and the activation function as in conventional DANNs, but \del{as well} \rw{also} through the desired polynomial degree of non-linearity per layer. Figure \ref{fig:dapc} schematically illustrates the neural processing in the DaPC NN through a multi-variate orthonormal basis $\Psi$ and a corresponding high-order weighted superposition $\sum\limits_{\forall d}$. Similar to the conventional DANN, unknown weights of the DaPC NN should be determined by training. 

In this sense, the paper does not aim to contribute to the DANN architecture, to any particular definition of loss functions, or to optimal training schemes. Instead, we offer a mathematical reformulation of the kernel DANN structure in terms of homogeneous chaos theory.  Moreover, introducing the high-order weighted superposition on each node may remove the necessity to introduce non-linearity via activation functions if desired. Hence, one can reduce the room for potential subjectivity in the modeling procedure.  We expect that joining the multi-layered structure of neural networks together with the theory of polynomial chaos expansion could be beneficial for ML tasks and also creates a foundation for further investigations.

The rest of the paper is structured as follows: Section~\ref{ML} summarizes the necessary background on the data-driven aPC in Section~\ref{ML_aPC} and the conventional DANN structure in Section~\ref{ML_DANN}. Section~\ref{ML_DANN} also points out the implicit Gaussian assumptions of neural signal processing in conventional DANN structures from the view point of PCE theory. Section~\ref{ML_DaPC} offers our novel DaPC NN formulation. It introduces the adaptive, deep, multi-variate orthonormal polynomial representation via a multi-layered structure. Section~\ref{ML_DaPC} also provides relevant properties that the DaPC NN inherits \del{form} \rw{from} aPC. Section~\ref{Section_DaPC_training} briefly outlines an example of the conventional training procedure to compute the weights by providing the required partial derivatives. Section \ref{Results} illustrates the performance of DaPC NN together with a conventional DANN and the aPC. In this comparison, we use exactly the same training data and loss function for DANN, DaPC NN and aPC in a total of three different test cases. 
Additionally, Section \ref{Results} shows evidence of convergence against reference validation data sets, \rw{comparing } \del{compering} how the size of the training data sets affects the performance of the considered ML approaches.

\section{Machine learning with non-redundant decomposition}
\label{ML}

\subsection{Arbitrary Polynomial Chaos Expansion}
\label{ML_aPC}
Theory of polynomial chaos expansion (PCE) was originally introduced by Norbert Wiener \cite{Wiener1938} in 1938. In PCE, the dependence of the model response (output) on all inputs is expressed using projection onto an orthogonal or orthonormal multi-variate polynomial basis \cite{Ghanem_Spanos_1991_SFEM_book,MR3364576}. For the current paper, we will employ a purely data-driven generalization of the PCE introduced in 2012 by Oladyshkin and Nowak \cite{OladNowak_RESS2012} that will open the pathway for a data-driven deep ML structure. 

\subsubsection{Homogeneus chaos theory}
\label{ML_aPC_chaos}
Let us consider a model response $\mathcal{R}(\bm{\omega})$ that depends on some multi-dimensional input $\bm{\omega}=\left\{\omega_1,\ldots,\omega_n \right\}$ from the input space $\Omega$, where $n$ is the number of inputs.
\rw{Let $L^2(\Omega)$ denote the $L^2$-space on $\Omega$, weighted by the assumed probability distribution.}
According to PCE theory \cite{MR0020230,Wiener1938}, the model response $\mathcal{R}(\bm{\omega})\in L^2(\Omega)$ can be expanded in $\bm{\omega}$ in the following manner to map $\mathbb{R}^n$ to $\mathbb{R}$:
\begin{eqnarray}
\label{PCE}
  \displaystyle \mathcal{R}(\bm{\omega}) = \sum_{i=0}^{\infty} w_{i} \Psi_i(\bm{\omega}) \approx \sum_{i=0}^{M} w_{i} \Psi_i(\bm{\omega}),
\end{eqnarray}
where $\Psi_i(\bm{\omega})$ are basis functions from a multi-variate orthonormal 
\rw{(w.r.t. the assumed probability distribution)}
polynomial basis $\big\{ \Psi_0(\bm{\omega}),\ldots, \Psi_M(\bm{\omega})\big\}$ defined on the input space $\Omega$, and $w_{i}$ are corresponding  coefficients that determine the form of the expansion in equation (\ref{PCE}). The total number of expansion terms $M$ depends on the number of inputs $N$ and on the desired degree $d$ of the polynomial representation as $M = (n+d)!/(n!d!)$. This formulation of $M$ is based on a total-degree truncation, and other alternatives exist\rw{~\cite{MR3364576}}. For a comprehensive discussion of the requirements for existence and completeness
of an orthonormal polynomial basis $\big\{\Psi_0(\bm{\omega}),\ldots,\Psi_M(\bm{\omega})\big\}$ of $L^2(\Omega)$,
we refer to \cite{MR2855645,MR2061539,MR3364576}.

The multi-variate polynomial basis $\big\{ \Psi_0(\bm{\omega}),\ldots, \Psi_M(\bm{\omega})\big\}$ is comprised of the tensor product of univariate orthonormal polynomials $\big\{\phi^{(0)}_j$, $\ldots$, $\phi^{(d)}_j \big\}$ of degree $d$ for the inputs $\omega_j$, assuming that the inputs are statistically independent:
\begin{eqnarray}
\label{basis_multi}
\Psi_{\alpha}(\bm{\omega})=\prod^N_{j=1} \phi^{(\alpha_j)}_j(\omega_j), \quad \sum^N_{j=1} \alpha_j \leq d.
\end{eqnarray}
Here $\alpha\rw{=(\alpha_1,\ldots,\alpha_N)\in\mathbb{N}^N_0}$ is a multivariate index \rw{describing polynomial degree} that contains the combinatoric information how to enumerate all possible products of individual univariate basis functions and contains the corresponding polynomial degree for input $\omega_j$ within the univariate polynomials $\phi^{(\alpha_j)}_j(\omega_j)$. The set of polynomials 
\del{$\big\{\phi^{(0)}_j$, $\ldots$, $\phi^{(d)}_j \big\}$}
\rw{$\big\{\phi^{(\alpha_j)}_j\mid \alpha_j=0, \ldots, d \big\}$}
forms an orthonormal basis of polynomial degree \rw{at most} $d$ for each input $\omega_j$:
\begin{eqnarray}
\label{multi_inde}
\left\langle \phi^{\del{p}\rw{(\alpha_j)}}_{j}(\omega_j), \phi^{\del{q}\rw{(\alpha_j')}}_j(\omega_j) \right\rangle_{L^2(\Omega)}= \delta_{\del{p,q}\rw{\alpha_j,\alpha_j'}},
\end{eqnarray}
where $\delta_{\del{p,q}\rw{\alpha_j,\alpha_j'}}$ represents the Kronecker delta $\forall \del{p,q}\rw{\alpha_j,\alpha_j'} =0,\ldots,d$.

\subsubsection{Data-driven orthonormal representation}
\label{ML_aPC_datadriven}
The original theory of homogeneous chaos \cite{Wiener1938} is based on orthonormal Hermite polynomials  
$\phi^{(\alpha_j)}_j(\omega_j)$ satisfying eq.~\eqref{multi_inde}, which are optimal \cite{OladNowak_RESS2012,MR3364576} for Gaussian distributed inputs $\bm{\omega}$. Further extensions to a  number of parametric statistical distributions (Gamma, Beta, Uniform, etc.) have been suggested in \cite{Xiu2002} and \cite{Xiu2003}, based on the Askey scheme \cite{Askey1985} of \rw{orthonormal} polynomials. 

Such approaches assume an exact knowledge of the involved probability density functions \cite{MR3364576}, which is not available in various applications or often requires additional assumptions \cite{RedHorse2004}\del{)}. To assure a purely data-driven representation for ML, in the current paper we will consider the data-driven generalization of the polynomial chaos expansion known as the arbitrary polynomial chaos (aPC) introduced in \cite{OladNowak_RESS2012}.  The necessity to adapt to arbitrary distributions in practical tasks is discussed in more detail in \cite{OladNowak_AWR2010}. 

The aPC technique adapts to arbitrary probability distribution shapes of the inputs and can be inferred from limited input data through a few statistical moments \cite{OladNowak_RESS2012}. Formally, univariate orthonormal polynomials bases $\phi^{(\alpha_j)}_j(\omega_j)$ of polynomial degree $\alpha_j$ \del{could} \rw{can} be written as a sum of the following monomials:
\begin{eqnarray}
\label{poly_monomials}
  \phi^{(\alpha_j)}_j(\omega_j) = \frac{1}{\sqrt{\kappa_{\alpha_j}}}\sum_{i=0}^{\alpha_j} m^{(\alpha_j)}_i \omega_j^i, \quad \alpha_j=\overline{0,d},
\end{eqnarray}
where $\kappa_{\alpha_j}=m^{(\alpha_j)}_{\alpha_j}$ is a constant representing the norm of the univariate polynomial. The corresponding monomial coefficients $m^{(\alpha_j)}_i$ \del{could} \rw{can} be defined according to the (empirical or theoretical) raw moments of the inputs $\omega_j$ as \del{following} \rw{follows} \cite{OladNowak_RESS2012,oladyshkin2018incomplete}:
\begin{eqnarray}
\label{orth_matrix_dd}     
\begingroup
\setlength\arraycolsep{1.8pt}
\begin{bmatrix}   
\mu_0(\omega_j) & \mu_1(\omega_j) & \ldots & \mu_{\alpha_j}(\omega_j) \cr                
\vdots & \vdots & \ddots & \vdots \cr            
\mu_{\alpha_j-1}(\omega_j) & \mu_{\alpha_j}(\omega_j) & \ldots & \mu_{2\alpha_j-1}(\omega_j) \cr
\mu_{\alpha_j}(\omega_j) & \mu_{\alpha_j+1}(\omega_j) & \ldots & \mu_{2\alpha_j}(\omega_j) \cr
\end{bmatrix} 
\endgroup
\begin{bmatrix}      
m^{(\alpha_j)}_0 \cr                
\vdots \cr            
m^{(\alpha_j)}_{\alpha_j-1} \cr                
m^{(\alpha_j)}_{\alpha_j} \cr    
\end{bmatrix} 
=
\begin{bmatrix}
0 \cr                
\vdots \cr            
0 \cr                
1 \cr    
\end{bmatrix},
\end{eqnarray}
where $\mu_{k}(\omega_j)$ denotes the $k$-th raw stochastic moment of the input $\omega_j$. 

Therefore, constructing the multi-variate orthonormal polynomial basis $\big\{ \Psi_0(\bm{\omega}),\ldots, \Psi_M(\bm{\omega})\big\}$ of degree $d$ for an aPC representation  $\mathcal{R}(\bm{\omega})$ in equation (\ref{PCE}) is based on the data-driven formulation in equations (\ref{basis_multi}), (\ref{poly_monomials}) and (\ref{orth_matrix_dd}). In particular, the aPC approach is based on $2d$ raw stochastic moments only. These moments \del{could} \rw{can} be evaluated directly from an available data set of limited size. \del{Additionally, the matrix} \rw{Matrix} on the left-hand side of equation (\ref{orth_matrix_dd}) is known as the Hankel matrix of moments \cite{Karlin1968} and its properties have been analysed in \cite{Lindsay1989}. The resulting polynomials are real if, and only if, the Hankel matrix of moments is positive definite, see also the related Hamburger moment problem, e.g. \cite{akhiezer1965classical,MR2855645,shohat1943problem,MR3364576}. From the practical point of view, the solution of the linear system of equations in (\ref{orth_matrix_dd}) \del{could} \rw{can} be obtained directly numerically, via lower-order moments representation \cite{OladNowak_RESS2012}, via recursive relations \cite{Abramowitz1965}, via Gram-Schmidt orthogonalization \cite{Witteveen2007} or via the Stieltjes procedure \cite{Stieltjes1884}.

Overall, the polynomial representation in equation (\ref{PCE}) is one of the oldest ML approaches to map $\mathbb{R}^n$ to $\mathbb{R}$ . It quantifies the response $\mathcal{R}$ to the inputs $\bm{\omega}$ through an orthonormal basis $\big\{ \Psi_0(\bm{\omega}),\ldots, \Psi_M(\bm{\omega})\big\}$ and computes the expansion coefficients $w_{i}$ \footnote{Traditionally, the response $\mathcal{R}$ has often been used as an approximation of some full-complexity physical model $\mathcal{M}$ in order to learn about the non-linear dependence of model output on modelling inputs $\bm{\omega}$, i.e. $\mathcal{R}(\bm{\omega}) \approx \mathcal{M}(\bm{\omega})$. In that sense the PCE projection in equation (\ref{PCE}) is often used a surrogate (response surface or reduced model), that is considered to be a special case of supervised machine learning.}. Each coefficient $w_{i}$ for $i>0$ indicates how much variance one or another term brings into the overall composition (see also relation to global sensitivity analysis in \cite{oladyshkin2012global,zhang2016evaluation}). The unknown expansion coefficients $w_{i}$ can be determined using Galerkin projection \cite{koppel2019comparison,OladNowak_RESS2012}, numerical integration, regression or collocation approaches \cite{Li2007,Olad_al_CG2009,Villadsen1978}.

\subsection{Deep Artificial Neural Networks}
\label{ML_DANN}

Let us shortly summarize the key aspects of conventional DANN structures. They all rely on  the multi-layer concept introduced by Alexey Ivakhnenko and Valentin Lapa in 1967 \cite{ivakhnenko1967cybernetics}. Similar to \rw{the method introduced in} Section \ref{ML_aPC}, DANNs maps $\mathbb{R}^n$ to $\mathbb{R}$ by providing an approximation of the response (output). 
\rw{In DANNs, information flows from input nodes, through co-called hidden layers, and then to output \cite{Goodfellow2016deeplearning}. The term hidden refers to the fact that the computations that occur within the hidden layer are not visible from the outside of the network. Each hidden layer in a neural network contains several hidden nodes, also known as hidden neurons. These nodes are responsible for performing computations on the input data and transmitting the results to other nodes in the network, ultimately leading to the output.}
It has gained significant popularity recently because it is a universal approximator \cite{hornik1989universal} and due to technological advances in computing power.

\subsubsection{Conventional DANN structure}

We will consider a conventional Deep Artificial Neural Network (DANN) with $L$ hidden layers and corresponding numbers $N^{(\mathcal{L})}$ of hidden neurons for each layer $\mathcal{L}$ ($\mathcal{L}=1,\ldots,L$). Similar to the ML approach in Section \ref{ML_aPC}, DANNs provide the response $\mathcal{R}(\bm{\omega})$ as a non-linear dependence \cite{anthony2009neural} on the same multidimensional input $\bm{\omega}=\left\{\omega_1,\ldots,\omega_n \right\}$ from the input space $\Omega$, where $n$ is the number of inputs. We will denote the response on each hidden layer $\mathcal{L}$ as a vector $\bm{\mathcal{R}}^{(\mathcal{L})}=\left\{\mathcal{R}^{(\mathcal{L},1)},\ldots,\mathcal{R}^{(\mathcal{L},N^{\mathcal{L}})}\right\}$. This vector represents the responses (outputs) from the corresponding hidden node $1,\ldots,N^{\mathcal{L}}$ on the current layer $\mathcal{L}$. Then, the response of the DANN representation $\mathcal{R}(\bm{\omega})$ could be \del{written} \rw{seen} \cite{Goodfellow2016deeplearning} as recursive encapsulation of responses \del{form} \rw{from} \rw{hidden} \del{deep} layers \rw{that contains $N^{(\mathcal{L})}$ hidden neurons  for each layer $\mathcal{L}$}. \rw{Formally, representation $\mathcal{R}(\bm{\omega})$} \del{Shortly} could be written in the following form: 
\begin{eqnarray}
\label{DANN_Response}
\mathcal{R}(\bm{\omega})=\bm{\mathcal{R}}^{(L)} \left(
\bm{\mathcal{R}}^{(L-1)}\left(\ldots\left(\bm{\mathcal{R}}^{(1)}\left(\bm{\omega}\right)\right)\right)\right),
\end{eqnarray}
where the input for each hidden layer $\bm{\mathcal{R}}^{(\mathcal{L})}$ is the response from the previous layer $\bm{\mathcal{R}}^{(\mathcal{L}-1)}$ ($\mathcal{L}=2,\ldots,L$). The input for the first layer $\bm{\mathcal{R}}^{(1)}$ is the overall input $\bm{\omega}=\left\{\omega_1,\ldots,\omega_n \right\}$ 

Each response $\mathcal{R}^{(\mathcal{L},\mathcal{N})}$ of the node $\mathcal{N}$ in the hidden layer $\mathcal{L}$ is defined according to the ANN representation \cite{Goodfellow2016deeplearning} as follows:
%
%
%
\begin{eqnarray}
\label{DANN_NodeResponse}
\begin{aligned}
\mathcal{R}^{(\mathcal{L},\mathcal{N})}\left(\bm{\mathcal{R}}^{(\mathcal{L}-1)}\right) =  &w^{(\mathcal{L},\mathcal{N})}_{0} \\
&+ \sum_{i=1}^{M^{(\mathcal{L})}} w^{(\mathcal{L},\mathcal{N})}_{i} \mathcal{A}^{(\mathcal{L})}(\mathcal{R}^{(\mathcal{L}-1,i)}).
\end{aligned}
\end{eqnarray}
Here, \rw{$\bm{\mathcal{R}}^{(\mathcal{L}-1)}=\left\{\mathcal{R}^{(\mathcal{L}-1,1)},\ldots,\mathcal{R}^{(\mathcal{L}-1,N^{\mathcal{L}-1})}\right\}$} is the response from the previous layer $\mathcal{L}-1$ (where $\bm{\mathcal{R}}^{(1)} =\bm{\omega}$), $M^{(\mathcal{L})}$ is the number of weights for node $\mathcal{N}^{(\mathcal{L}-1)}$, $\mathcal{A}^{(\mathcal{L})}$ is \del{an} \rw{the} activation function\footnote{The activation function could be also specified for each individual node $\mathcal{N}$ in the layer $\mathcal{L}$ as $\mathcal{A}^{(\mathcal{L},\mathcal{N})}$, but in order to keep transparency for the reader we keep the formulation where the activation function $\mathcal{A}^{(\mathcal{L})}$ is the same for all nodes of the layer} for the layer $\mathcal{L}$ , $w^{(\mathcal{L},\mathcal{N})}_{0}$ is \del{a} \rw{the} bias in the node $\mathcal{N}$ of the layer $\mathcal{L}$ and $w^{(\mathcal{L},\mathcal{N})}_{i}$ ($i=1,\ldots,M^{(\mathcal{L})}$) are the weights in the node $\mathcal{N}$ of layer $\mathcal{L}$. \rw{The notation $\mathcal{R}^{(\mathcal{L},\mathcal{N})}$ in equation (\ref{DANN_NodeResponse}) represents the non-activated response in the current paper \del{(e.g. as the overall DANN response constructed on the last layer)} and, hence, $\mathcal{A}^{(\mathcal{L})}(\mathcal{R}^{(\mathcal{L},\mathcal{N})})$ corresponds to the activated response.} \rw{We would like to clarify to the reader, that there are two common ways \cite{anthony2009neural} to write the equation for the response on a hidden node of a DANN using post and pre-activation function. The post-activation function applies directly to the inputs from the previous layer before computing the weighted sum (as in equation (\ref{DANN_NodeResponse})), while the pre-activation function applies to the weighted sum of inputs from the previous layer. However, once the pre-activation formulation is utilized, the output of DANN is generated without use of an activation function. Despite these two different ways of writing the equation, they ultimately lead to the same formal representation of the response as a function of the inputs. This is because the two formulations are mathematically equivalent and can be transformed into each other using simple algebraic manipulations.}

There is a variety of non-linear (and also linear) functions that \del{can be} \rw{are commonly} applied as activation functions $\mathcal{A}^{(\mathcal{L})}$ for an arbitrary input $\mathcal{I}$. The most popular activation functions choice are \rw{the} sigmoid in equation (\ref{DANN_Sigmoid}), \rw{the} hyperbolic tangent in equation (\ref{DANN_Tanh}) and \rw{the} rectified linear unit in equation (\ref{DANN_ReLU}):
\begin{eqnarray}
\label{DANN_Sigmoid}
\mathcal{A}_{sig}^{(\mathcal{L})}(\mathcal{I}) =  \frac{1}{1+e^{-\mathcal{I}}}.
\end{eqnarray}
\begin{eqnarray}
\label{DANN_Tanh}
\mathcal{A}_{tanh}^{(\mathcal{L})}(\mathcal{I})  =  \frac{e^{\mathcal{I}} - e^{-\mathcal{I}}}{e^{\mathcal{I}} + e^{-\mathcal{I}}}.
\end{eqnarray}
\begin{eqnarray}
\label{DANN_ReLU}
\mathcal{A}_{ReLU}^{(\mathcal{L})}(\mathcal{I})  =  \max(\mathcal{I},0).
\end{eqnarray}
 The decision to choose a specific function depends heavily on the prediction task and the data type \cite{sharma2017activation}. Furthermore, care has to be taken when choosing the activation functions, as it can lead to \rw{the so-called} vanishing or exploding gradient problems during DANN training \cite{Aggarwal2018neuralnetworks}.

The weights \del{could  be seeing} \rw{can be seen} as a vector
$\bm{w}=\big\{w^{(\mathcal{L},\mathcal{N})}_{i},i=1,\ldots,N_w \big\}$ with the total number of weights $N_w$ that depends on the number of layers $L$, the number of hidden neurons $N^{(\mathcal{L})}$ per layer $\mathcal{L}$ ($\mathcal{L}=1,\ldots,L$) and the number of weights $M^{(\mathcal{L})}$ per hidden neurons $N^{(\mathcal{L})}$ of the layer $\mathcal{L}$. The weights $w^{(\mathcal{L},\mathcal{N})}_{i}$ of the DANN define the final form of the representation $\mathcal{R}(\bm{\omega})$ in equation (\ref{DANN_Response}), and they are determined via a training procedure. More details of the training procedure are discussed in Section \ref{Section_DaPC_training}.

\subsubsection{Connection between DANN and aPC}
\label{DANNmonomials}
According to equation (\ref{DANN_NodeResponse}), the conventional structure of DANNs propagates the signal from inputs to the response through deep layers, where each node of the layer employs a linear combination of \rw{zeroth and first-order} monomials. The monomials represent the activated output from the nodes of the previous layer. Considering the definitions in Section \ref{ML_aPC}, the linear representation via monomials is a particular case of the PCE theory, where the polynomial basis of $0^{th}$ degree $\phi^{(0)}_j$ and $1^{st}$ degree $\phi^{(1)}_j$ is defined explicitly as:
\begin{eqnarray}
\label{poly_linear_monomials}
\phi^{(0)}_j(\omega_*) = m^{(0)}_0, \quad \phi^{(1)}_j(\omega_*) = m^{(1)}_0+m^{(1)}_1 \omega_i,
\end{eqnarray}
where $\omega_i$ is some input of a hidden node $\mathcal{N}$ in a layer $\mathcal{L}$. 

\rw{As it have been stated in Section \ref{ML_aPC_chaos}, the PCE theory requires that the set of polynomials (as well in equation \ref{poly_linear_monomials})) forms an orthonormal basis for each input $\omega_j$ in the input space $\Omega$, i.e. satisfying equation (\ref{multi_inde}). For example, the original theory of homogeneous chaos \cite{Wiener1938} is based on Hermite polynomials that are orthonormal for Gaussian distributed inputs. The use of any other basses for Gaussian distributed inputs will result in an erroneous non-orthogonal decomposition and considered to be not optimal \rw{\cite{OladNowak_RESS2012,MR3364576}}}. The aPC theory \cite{OladNowak_RESS2012} \rw{generalizes the original PCE theory by allowing for arbitrary input distributions and} constructs the orthonormal polynomial basis from the available data-driven input distribution \rw{that is encoded in raw moments}. \rw{However, the conventional DANN imposes the basis via a particular form of $0^{th}$ and $1^{st}$ order polynomial in equation (\ref{poly_linear_monomials})}, but we do not know for which underlying distribution that polynomial basis would satisfy the orthonormality conditions, i.e. whether it is optimal. To find out the  underlying \rw{orthonormal} distribution \rw{for the conventional DANN that satisfies equation (\ref{multi_inde})}, we will explore equations (18) and (22) from the paper \cite{oladyshkin2018incomplete}\rw{, that are written as following:
\begin{eqnarray}
\begin{bmatrix}
m_0^{(0)} & 0 & \cdots & 0 \\
m_0^{(1)} & m_1^{(1)} & \cdots & 0 \\
\vdots & \vdots & \ddots & \vdots \\
m_0^{(\alpha_j)} & m_1^{(\alpha_j)} &\cdots & m_{\alpha_j}^{(\alpha_j)}
\end{bmatrix}
\begin{bmatrix}
\mu_0 \\ \mu_1 \\ \vdots \\ \mu_{\alpha_j}
\end{bmatrix}
=
\begin{bmatrix}
1\\ 0 \\ \vdots \\ 0
\end{bmatrix} \nonumber
\end{eqnarray}
and 
\begin{eqnarray}
\begin{bmatrix}
m_0^{(0)} & 0 & \cdots & 0 \\
m_0^{(1)} & m_1^{(1)} & \cdots & 0 \\
\vdots & \vdots & \ddots & \vdots \\
m_0^{(\alpha_j)} & m_1^{(\alpha_j)} &\cdots & m_{\alpha_j}^{(\alpha_j)}
\end{bmatrix}
\begin{bmatrix}
\mu_1 \\ \mu_2 \\ \vdots \\ \mu_{\alpha_j+1}
\end{bmatrix}
=
\begin{bmatrix}
m_0^{(0)}\mu_1\\ 1 \\  \vdots \\ 0
\end{bmatrix}.\nonumber
\end{eqnarray}
}
\rw{Solving the system of linear equations presented above, we can} reconstruct the first two raw moments behind the polynomial basis in equation (\ref{poly_linear_monomials}) \rw{that are used in the conventional DANN. This process entails}:
\begin{eqnarray}
\label{DANN_monomials}
  \mu_1=0, \mu_2=1.
\end{eqnarray}
Therefore, we can conclude that conventional DANN structures implicitly assume a Gaussian distribution with zero mean and unit variance \rw{(i.e. standard Gaussian distribution)} for propagating the signal through hidden layers. In other words, the conventional DANN representation \rw{optimally preserves orthonormality}, if and only if, all neural inputs of all hidden layers were standard Gaussian. \rw{However, if the inputs of all hidden layers do not follow a standard Gaussian distribution, then the conventional DANN structure commonly used in machine learning may not be the most optimal way to propagate the signals through the hidden layers.} \del{Apparently, applied} \rw{Applied} ML tasks could often be used in situations where the propagation of neural signals \del{form} \rw{from} layer to layer is not necessarily Gaussian, let alone standardized to unit variance. Therefore, based on \rw{results presented in} \cite{OladNowak_RESS2012} and \cite{oladyshkin2018incomplete}, we \rw{conclude that } \del{consider} the linear representation via \rw{non-orthonormal} monomials \rw{is} \del{to be maximum} not optimal. 
Not even batch normalization \cite{ioffe2015batch} of each node's input could mitigate the mentioned effect due to its linear nature. Theoretically, some general non-linear transformation could map the distributions of all neural inputs onto the Gaussian, but this is not feasible due to the  data-driven nature of such neural inputs.

\subsection{Deep Arbitrary Polynomial Chaos Neural Network}
\label{ML_DaPC}

Let us generalize the structure of conventional DANNs in Section \ref{ML_DANN} to overcome their redundancy \del{and non-ortho-} \del{normality in} \rw{caused by the non-orthonormal} representation of neural signals. To do so, we will employ the orthogonal representation via the data-driven theory of polynomial chaos expansion introduced in Section \ref{ML_aPC}. This also introduces the possibility to consider additional high-order interactions between neurons through the non-linear \del{linear} \rw{multivariate} terms from the PCE representation in the conventional DANN structure in equation (\ref{DANN_NodeResponse}). 

\subsubsection{Deep orthonormal polynomial representation}
\label{Deep_OPR}
We will consider the number of deep layers $L$ and the corresponding number of \del{deep} neurons $N^{(\mathcal{L})}$ for each layer $\mathcal{L}$ as in Section \ref{ML_DANN}. Similar to Section \ref{ML_aPC} and Section \ref{ML_DANN}, we will map the multi-dimensional input $\bm{\omega}=\left\{\omega_1,\ldots,\omega_n \right\}$ to a response $\mathcal{R}(\bm{\omega})$. Combining the deep ML representation in equation (\ref{DANN_Response}) with the orthonormal expansion in equation (\ref{PCE}), we will construct a generalized representation denoted as Deep Arbitrary Polynomial Chaos Neural Network (DaPC NN). Here, the response $\mathcal{R}^{(\mathcal{L},\mathcal{N})}$ of the \del{deep} node $\mathcal{N}$ in the hidden layer $\mathcal{L}$ forms a vector of layer responses $\bm{\mathcal{R}}^{(\mathcal{L})}=\left\{\mathcal{R}^{(\mathcal{L},1)},\ldots,\mathcal{R}^{(\mathcal{L},N^{\mathcal{L})}}\right\}$ as follows: 
%
%
%
\begin{equation}
\label{DaPC_NodeResponse}
\begin{aligned}
\mathcal{R}^{(\mathcal{L},\mathcal{N})}\left(\bm{\mathcal{R}}^{(\mathcal{L}-1)}\right) = &  \sum_{i=0}^{M^{(\mathcal{L})}} w^{(\mathcal{L},\mathcal{N})}_{i} \Psi^{(\mathcal{L})}_i 
\Big[\mathcal{A}^{(\mathcal{L})}(\mathcal{R}^{(\mathcal{L}-1,1)}),\\  
& ...,\mathcal{A}^{(\mathcal{L})}(\mathcal{R}^{(\mathcal{L}-1,N^{\mathcal{L}-1})})
\Big],
\end{aligned}  
\end{equation}
where \rw{$\bm{\mathcal{R}}^{(\mathcal{L}-1)}=\left\{\mathcal{R}^{(\mathcal{L}-1,1)},\ldots,\mathcal{R}^{(\mathcal{L}-1,N^{\mathcal{L}-1})}\right\}$} is the neural response from the previous layer $\mathcal{L}-1$ (again with $\bm{\mathcal{R}}^{(1)} =\bm{\omega}$), $M^{(\mathcal{L})}$ is the total number of terms for each node on the layer $\mathcal{L}$, $\mathcal{A}^{(\mathcal{L})}$ is an activation function for the layer $\mathcal{L}$, $\Psi^{(\mathcal{L})}_i$ is a multivariate orthonormal \rw{(w.r.t. the probability distribution given by response of the previous layer)} polynomial from the basis $\big\{ \Psi^{(\mathcal{L})}_0,\ldots, \Psi^{(\mathcal{L})}_{M^{(\mathcal{L})}}\big\}$ of degree $d^{(\mathcal{L})}$ for the layer $\mathcal{L}$ and $w^{(\mathcal{L},\mathcal{N})}_{i}$ are the weights of the node $\mathcal{N}$ in layer $\mathcal{L}$, where the term $w^{(\mathcal{L},\mathcal{N})}_{0}$ represents bias as in the conventional DANN definition in equation (\ref{DANN_NodeResponse}).

 The DaPC NN representation in equation (\ref{DaPC_NodeResponse}) reflects the non-linear interaction between the neurons using high-order \rw{multivariate} terms and orthonormal representation in contrast to DANNs in equation (\ref{DANN_NodeResponse}). According to Section \ref{ML_aPC}, the total number of weights $M^{(\mathcal{L})}$ on each node $\mathcal{N}$ of the layer $\mathcal{L}$ depends on the number of layer inputs $N^{(\mathcal{L}-1)}$ from the previous layer and on the desired degree $d^{(\mathcal{L})}$ of the polynomial representation for the layer $\mathcal{L}$ as:
\begin{eqnarray}
\label{DaPC_NofW}
M^{(\mathcal{L})} = \frac{(N^{(\mathcal{L}-1)}+d^{(\mathcal{L})})!}{N^{({\mathcal{L}-1})}!d^{(\mathcal{L})}!}.
\end{eqnarray}
To ensure the optimal transfer of neural signals from layer to layer and to mitigate redundancy \rw{within each node} by the DaPC NN representation, we will construct the multivariate orthonormal polynomial basis $\big\{ \Psi^{(\mathcal{L})}_0,\ldots, \Psi^{(\mathcal{L})}_{M^{(\mathcal{L})}}\big\}$ of degree $d^{(\mathcal{L})}$ for each layer $\mathcal{L}$ depending on the layer input, i.e. depending on the activated response from the previous layer $\mathcal{A}^{(\mathcal{L})} (\bm{\mathcal{R}}^{(\mathcal{L}-1)})$. 

As the input layer ($\mathcal{L}=1$) directly corresponds to the inputs $\bm{\omega}$ (i.e. $\bm{\mathcal{R}}^{(1)}=\bm{\omega}$) and the inputs $\bm{\omega}$ follow some arbitrary (but given by each specific application) data-driven distribution, the response $\bm{\mathcal{R}}^{(1)}$ follows the exactly same distribution. For example, the input training data set could be employed in a purely data-driven way to serve as empirical distribution. After that, the responses $\bm{\mathcal{R}}^{(\mathcal{L})}$ ($\mathcal{L}=2,..,L$) on the all nodes of the multi-layered structure \rw{will} \del{could again} follow \del{any arbitrary data-driven} distributions that result from all previous weights, biases, polynomials and activation functions. That means, the procedure of orthonormalization proceeds sequentially (forward) through the layers. \del{To respect the data-driven structure of the orthonormalization procedure, we will employ the aPC representation to compute the data-driven multi-variate orthonormal basis introduced in equations (\ref{basis_multi}), (\ref{poly_monomials}) and (\ref{orth_matrix_dd}) to construct the corresponding orthonormal bases $\big\{ \Psi^{(\mathcal{L})}_0,\ldots, \Psi^{(\mathcal{L})}_{M^{(\mathcal{L})}}\big\}$ in each layer.}
\rw{To maintain the data-driven approach of the orthonormalization procedure, we will utilize the aPC representation for computing the data-driven multi-variate orthonormal basis. This basis is introduced in equations (\ref{basis_multi}), (\ref{poly_monomials}) and (\ref{orth_matrix_dd}), which will be used to construct the corresponding orthonormal bases $\big\{ \Psi^{(\mathcal{L})}_0,\ldots, \Psi^{(\mathcal{L})}_{M^{(\mathcal{L})}}\big\}$ in each layer.}
Automatically, by ensuring the orthogonal decomposition, the weights $w^{(\mathcal{L},\mathcal{N})}_{i}$ of a particular node $\mathcal{N}$ and a layer $\mathcal{L}$ gain a meaning according to  global sensitivity analysis \cite{oladyshkin2012global}: the weights reflects the partial contribution of each single neuron (linear \rw{univariate} terms) or simultaneous combination of neurons (non-linear \rw{multivariate} terms) to the total variance of the response $\mathcal{R}^{(\mathcal{L},\mathcal{N})}$ for the node $\mathcal{N}$ and the layer $\mathcal{L}$.

The training procedure itself will be discussed in Section \ref{Section_DaPC_training}, where the weights $\bm{w}=\left\{w^{(\mathcal{L},\mathcal{N})}_{i},i=1,\ldots,N_w \right\}$ in equation (\ref{DaPC_NodeResponse}) will be obtained via a similar training procedure as for DANNs.  However, independent from any particular training procedure, the weights $\bm{w}$ determine uniquely the corresponding data-driven orthonormal bases $\big\{ \Psi^{(\mathcal{L})}_0,\ldots, \Psi^{(\mathcal{L})}_{M^{(\mathcal{L})}}\big\}$ for each layer $\mathcal{L}$ ($\mathcal{L}=1,\ldots,L$) of DaPC NNs. Indeed, the orthonormal basis on a particular layers $\mathcal{L}$ depends on the neural response from the previous layer $\bm{\mathcal{R}}^{(\mathcal{L}-1)}$ only. Remarking that $\bm{\mathcal{R}}^{(1)} =\bm{\omega}$, the corresponding data-driven orthonormal bases $\big\{ \Psi^{(2)}_0,\ldots, \Psi^{(2)}_{M^{(2)}}\big\}$ for the second layer (i.e. $\mathcal{L}=2$) \del{could} \rw{can} be constructed via equations (\ref{basis_multi}), (\ref{poly_monomials}) and (\ref{orth_matrix_dd}). The neural response  $\bm{\mathcal{R}}^{(2)}$ on the second layer itself is again fully determined by the corresponding weights through equation (\ref{DaPC_NodeResponse}). Taking into consideration the recursive encapsulation of neural responses in equation (\ref{DANN_Response}), it is easy to see that orthonormal bases on all layers are dictated by the weights only. Therefore, all trained weights $\bm{w}$ determine uniquely the corresponding data-driven orthonormal bases $\big\{ \Psi^{(\mathcal{L})}_0,\ldots, \Psi^{(\mathcal{L})}_{M^{(\mathcal{L})}}\big\}$ for each layer $\mathcal{L}$ ($\mathcal{L}=1,\ldots,L$) of DaPC NN without any additionally actions.

\begin{figure*}[!t]
	\centering
	\includegraphics[width=1.0\linewidth]{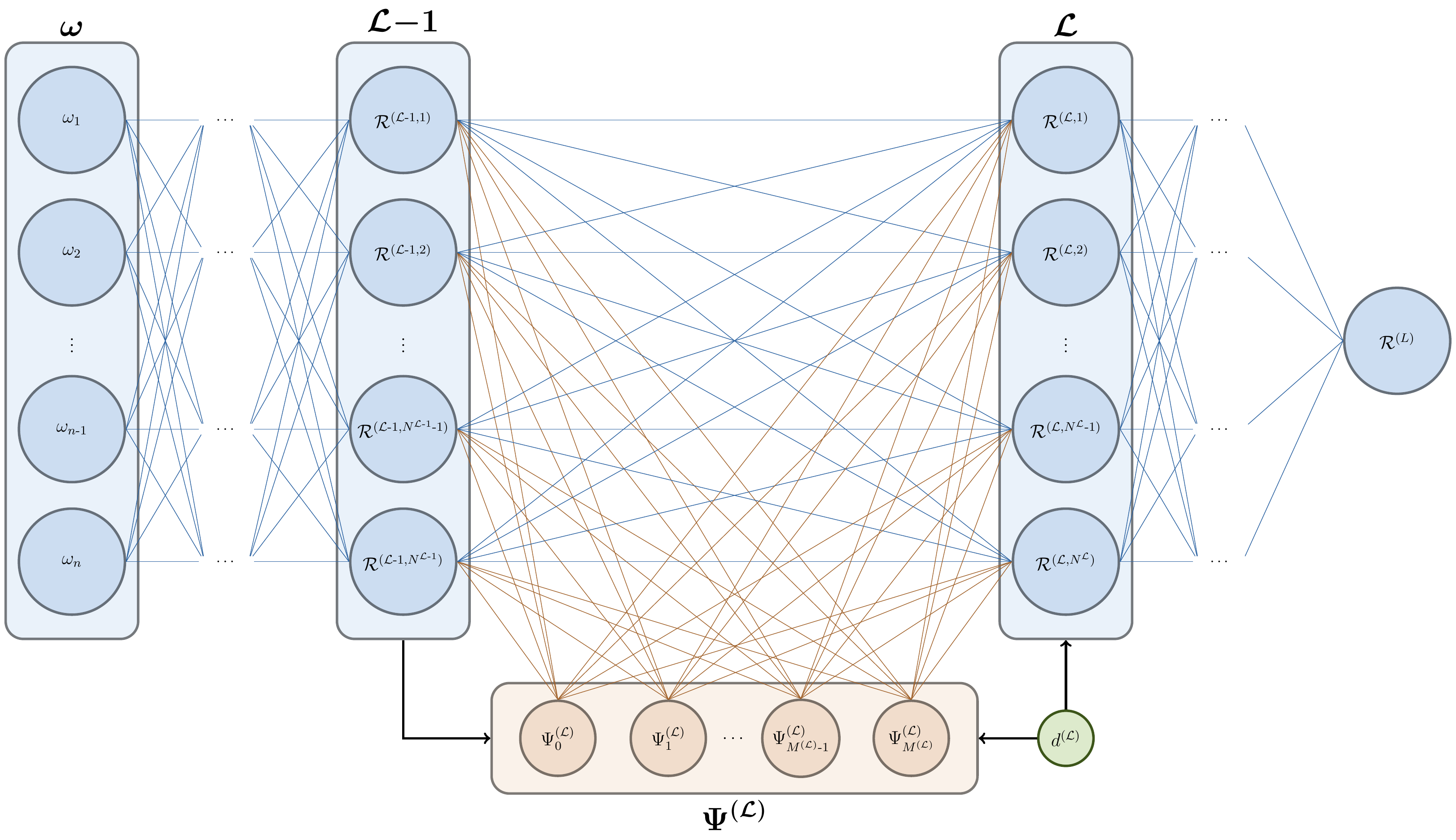}
	\caption{An illustration of the DaPC NN structure. The input of layer $\mathcal{L}$ is formed by the output of layer $\mathcal{L}-1$ and the orthonormal basis $\big\{ \Psi^{(\mathcal{L})}_0,\ldots, \Psi^{(\mathcal{L})}_{M^{(\mathcal{L})}}\big\}$ of the polynomial degree $d^{(\mathcal{L})}$.}
 \label{fig:model}
\end{figure*}

Figure \ref{fig:model} schematically illustrates the structure of the introduced DaPC NN.  Similar to DANNs, the structure of DaPC NNs is specified via hidden layers $\mathcal{L}$ ($\mathcal{L}=1,\ldots,L$), hidden nodes $\mathcal{N}$ ($\mathcal{N}=1,\ldots,N^{(\mathcal{L})}$) and activation functions $\mathcal{A}^\mathcal{L}$. 
\del{However, each} \rw{Each} layer $\mathcal{L}$ is equipped with \del{the} \rw{an} orthonormal basis  $\big\{ \Psi^{(\mathcal{L})}_0,\ldots, \Psi^{(\mathcal{L})}_{M^{(\mathcal{L})}}\big\}$.
Basically, neural signals traveling from layer to layer should pass through a sort of data-driven filter that \del{allocates} \rw{constructs} an optimal orthonormal representation for each layer \del{additively} as illustrated in Figure \ref{fig:model}. Such an orthonormal basis is the same for all nodes $\mathcal{N}$ ($1,\ldots,N^{(\mathcal{L})}$) of a given layer $\mathcal{L}$ and employs the neural signal coming as response $\mathcal{R}^{(\mathcal{L}-1,\mathcal{N})}$ from the previous layer $\mathcal{L}-1$.  Moreover, the suggested DaPC NN structure offers flexibility to specify the degree of non-linearity $d^{(\mathcal{L})}$ for each particular layer $\mathcal{L}$ ($\mathcal{L}=1,\ldots,L$) in order to go beyond a linear representation of \rw{univariate neurons}, if desired.
 
 \subsubsection{Properties of the DaPC NN}
 \label{DaPC_prop}
 
Let us introduce relevant properties of the DaPC NN that could be useful for practical applications.

\paragraph{Property 1:} Due to the orthonormal DaPC NN representation, the expected value $\mu \left(\mathcal{R}^{(\mathcal{L},\mathcal{N})}\right)$ and total variance $\sigma^2\left(\mathcal{R}^{(\mathcal{L},\mathcal{N})}\right)$ of the response $\mathcal{R}^{(\mathcal{L},\mathcal{N})}$ on each particular node $\mathcal{N}$ of any layer $\mathcal{L}$ (including the last layer forming the total response $\mathcal{R}$) can be quantified analytically \cite{OladNowak_RESS2012} using the following explicit form:
\begin{eqnarray}
\label{DaPC_meanvar}
& \mu \left(\mathcal{R}^{(\mathcal{L},\mathcal{N})}\right)=w^{(\mathcal{L},\mathcal{N})}_{0},\\
&\sigma^2\left(\mathcal{R}^{(\mathcal{L},\mathcal{N})}\right)=\sum_{i=1}^{M^{(\mathcal{L})}}\left(w^{(\mathcal{L},\mathcal{N})}_{i}\right)^2,
\end{eqnarray}
where the explicit analytical relations in equation (\ref{DaPC_meanvar}) are written with respect to the response $\bm{\mathcal{R}}^{(\mathcal{L}-1)}$ from the previous layer $\mathcal{L}$ and not with respect to the inputs $\bm{\omega}$.

\paragraph{Property 2:}
Due to the orthonormal DaPC NN representation, Sobol \cite{sobol1990sensitivity} sensitivity indices  $S^{(\mathcal{L},\mathcal{N})}_{I}$ of a particular node $\mathcal{N}$ and layer $\mathcal{L}$ can be explicitly computed as follows according to global sensitivity analysis \rw{\cite{oladyshkin2012global,SUDRET2008964}}: 
\begin{equation}
 S^{(\mathcal{L},\mathcal{N})}_{I}= \frac{\left(w^{(\mathcal{L},\mathcal{N})}_{I}\right)^2}{\sum_{i=1}^{M^{(\mathcal{L})}}\left(w^{(\mathcal{L},\mathcal{N})}_{i}\right)^2},
\end{equation}
where the Sobol index $S^{(\mathcal{L},\mathcal{N})}_{I}$ reflects the relative partial contribution of each single neuron (linear \rw{univariate} terms) or simultaneous combination of neurons (non-linear \rw{multivariate} terms) to the total variance of the response $\mathcal{R}^{(\mathcal{L},\mathcal{N})}$ for that particular node $\mathcal{N}$ in a layer $\mathcal{L}$.

\paragraph{Property 3:}  An analytical form for the partial derivatives of the DaPC NN representation with respect to each particular weight can be obtained by applying the chain rule of differentiation \cite{Goodfellow2016deeplearning}. Considering the recursive encapsulation in equation (\ref{DANN_Response}) and the orthonormal representation in equation (\ref{DaPC_NodeResponse}), we obtain the following analytical form for the partial derivative \footnote{Partial derivatives are functions of the inputs $\bm{\omega}$} of the  response $\mathcal{R}^{(\mathcal{L},\mathcal{N})}$ in the node $\mathcal{N}$ of the layer $\mathcal{L}$ with respect to a weight $w^{(\mathcal{L_*},\mathcal{N_*})}_{k}$ ($\forall k \in [1,M^{(\mathcal{L_*})}]$) in the node $\mathcal{N_*}$ of the layer $\mathcal{L_*}$ when $\mathcal{L}>\mathcal{L_*}$:
\begin{equation}
\label{DaPC_derivatives}
\begin{aligned}
\displaystyle \frac{\partial  \mathcal{R}^{(\mathcal{L},\mathcal{N})}}{\partial w^{(\mathcal{L_*},\mathcal{N_*})}_{k}} = &\sum_{i=0}^{M^{(\mathcal{L})}} w^{(\mathcal{L},\mathcal{N})}_{i} 
\sum_{j=0}^{N^{(\mathcal{L}-1)}}
\Bigg(\frac{\partial \Psi^{(\mathcal{L})}_i}{\partial \mathcal{A}^{(\mathcal{L})}}\cdot 
\\
&\cdot \frac{\partial \mathcal{A}^{(\mathcal{L})}}{\partial  \mathcal{R}^{(\mathcal{L}-1,j)}}\frac{\partial \mathcal{R}^{(\mathcal{L}-1,j)}}{\partial w^{(\mathcal{L_*},\mathcal{N_*})}_{k}}\Bigg)
\end{aligned}
\end{equation}
and the following simplified analytical form when $\mathcal{L}=\mathcal{L_*}$:
%
%
\begin{equation}
\label{DaPC_derivatives_first}
\begin{aligned}
\displaystyle \frac{\partial  \mathcal{R}^{(\mathcal{L_*},\mathcal{N_*})}}{\partial w^{(\mathcal{L_*},\mathcal{N_*})}_{k}} =   \Psi^{(\mathcal{L_*})}_k\Big[& \mathcal{A}^{(\mathcal{L_*})}(\mathcal{R}^{(\mathcal{L_*}-1,1)}), \\ &...,\mathcal{A}^{(\mathcal{L_*})}(\mathcal{R}^{(\mathcal{L_*}-1,N^{\mathcal{L_*}-1})})
\Big],
\end{aligned}  
\end{equation}
where, similar to above, the same analytical relations for partial derivatives hold for the DaPC NN response $\mathcal{R}$ as the response of the last layer  $\bm{\mathcal{R}}^{(L)}$. 

\paragraph{Remark 1:} The data-driven aPC representation $\mathcal{R}_\mathrm{aPC}(\bm{\omega})$ in Section \ref{ML_aPC} is a particular case of a DaPC NN representation $\mathcal{R}_\mathrm{DaPCNN}(\bm{\omega})$ when the number of hidden layers is equal to one:
\begin{eqnarray}
\label{DaPC_aPC_link}
\mathcal{R}_\mathrm{aPC}(\bm{\omega})=\mathcal{R}_\mathrm{DaPC NN}(\bm{\omega}), \; \mathrm{for} \; \mathcal{L}=1. 
\end{eqnarray}

\paragraph{Remark 2:} The conventional DANN representation $\mathcal{R}_\mathrm{DANN}(\bm{\omega})$ in Section \ref{ML_DANN} is a particular case of the DaPC NN representation $\mathcal{R}_\mathrm{DaPCNN}(\bm{\omega})$ when the degree of expansion in each layer is equal to one and the basis is defined according to equation (\ref{poly_linear_monomials}) for a standard Gaussian distribution of neuronal signals in each hidden node: 
\begin{eqnarray}
\label{DaPC_aPC_link}
\mathcal{R}_\mathrm{DANN}(\bm{\omega})=\mathcal{R}_\mathrm{DaPC NN}(\bm{\omega}), \; \mathrm{for} \; d^\mathcal{L}=1, \forall\mathcal{L}=1,\ldots,L.
\end{eqnarray}

\paragraph{Remark 3:} 
The DaPC NN offers a possibility to reflect non-linearities of neural responses both by introducing the high-order expansion $d^{(\mathcal{L})}$ for the responses $\mathcal{R}^{(\mathcal{L},\mathcal{N})}$ of the hidden nodes and by introducing non-linearity in the activation functions $\mathcal{A}^{(\mathcal{L})}$. \rw{Theoretically, there are no restrictions on the use of activation functions within the current DaPC NN framework according to equation (\ref{DANN_NodeResponse}). The modeler has the freedom to determine an activation function if it would be advantageous for a specific modeling procedure since equations (\ref{DaPC_derivatives}) and (\ref{DaPC_derivatives_first}) provide the analytical form of the partial derivatives that include the activation function.} However, simultaneously introducing non-linearities both in expansion degrees and in activation functions could lead to difficulties in network training caused by strong numerical round-off errors. We would rather suggest to employ linear activation functions while shifting the representation of non-linear dependence to the orthonormal decomposition. Especially, any linear normalization (similar to batch normalization \cite{ioffe2015batch}) will mitigate potential numerical challenges during the training procedure. Therefore, in the current paper we suggest to employ the analytical relations introduced in \textit{Property 1} to have a handle on total (explained) variances for the overall network. This can be achieved via the following normalized activation function: 
%
\begin{eqnarray}
\label{DaPC_normalization}
\mathcal{A}^{(\mathcal{L})} (\mathcal{I})=\frac{\mathcal{I}-\mu\left(\mathcal{I}\right)}{\sigma\left(\mathcal{I}\right)}.
\end{eqnarray}

\subsection{Computation of weights}
\label{Section_DaPC_training}

\subsubsection{Training data}
The exact forms of aPC in Section \ref{ML_aPC}, DANN in Section \ref{ML_DANN} and DaPC NN in Section \ref{ML_DaPC} are determined via corresponding weights that are typically computed using training data in a training procedure. Let us denote the training inputs by $\bm{\omega}_T=\{\bm{\omega}_{T_1},\ldots,\bm{\omega}_{T_N} \}$ and the corresponding training responses by $\mathbf{R}_{T_N}=\{R_{T_1},\ldots,R_{T_N}\}$, where $T_N$ is a value greater than zero representing the size of training data sets $\bm{D}_{T_N}=\left\{(\bm{\omega}_{Ti},R_{Ti}), i=1,\ldots,T_N \right\}$. Such data could be obtained from runs of an original physical model for surrogate modelling \cite{OladNowak_RESS2012,beckers2020bayesian}, or directly from a given database for other learning tasks \cite{deng2009imagenet,krizhevsky2017cnn,huang2018densely}. In order to assure transparency in the current paper, we will consider a fixed training set of size $T_N$, avoiding any data mining procedure such as active learning \rw{\cite{oladyshkin2020bayesian3,settles.tr09}}. Thus, the data set $\bm{D}_{T_N}$ is the complete training set for the discussed ML approaches. 

\subsubsection{Training procedure}

Deep structures like conventional DANNs in Section \ref{ML_DANN} or the newly introduced DaPC NN in Section \ref{ML_DaPC} require solving a non-linear system of equations to obtain the unknown weights $w^{(\mathcal{L},\mathcal{N})}_{i}$. There are two possible ways to solve the corresponding system of non-linear equations. The first way is based on deterministic approaches, such as gradient-based search \cite{ruder2017overview} or the Levenberg-Marquardt algorithm \cite{marquardt1963} \rw{that have been widely used in deep learning}. \rw{In the current paper, we will follow such a popular way for training of DaPC NN to ensure that the training process is comparable to the traditional practices for conventional DANN structures.} However, \rw{for the sake of completeness we would like to mention, that} the underlying problem is ill-posed, so that a unique deterministic solution to the training problem may not exist \cite{Adler_2017,yee1993ill}. Therefore, the second way helps to solve the challenges related to ill-\rw{posedness }\del{possess} of the problem. It is based on stochastic inference \cite{kolmogorov2018foundations} by seeing the weights as random variables, and then conditioning the weights $w^{(\mathcal{L},\mathcal{N})}_{i}$ on the available training data \cite{praditia2022learning}.  However, straightforward stochastic approaches such as Monte Carlo \cite{Smith1992} or even Markov chain Monte Carlo \cite{Gilks1996} are computationally very expensive, and they suffer in cases with high non-linearity and high-dimensionality \cite{papamarkou2021challenges}. Recently, there are some works trying to overcome the computational problem of stochastic inference assuming Gaussianity and mutual independence of weights \cite{blundell2015bnn}.  Alternatively, combinations of deterministic approaches with a regularization helping to deal with overfitting \cite{Goodfellow2016deeplearning} have \del{showed reasonably acceptable results and have} been very popular in the last decades for DANNs.

Therefore, in the current paper, we will adopt \del{that popular way} \rw{the widely used approach, and extend the optimization procedure by Tikhonov or ridge regularization \cite{tikhonov1977solutions},} \rw{to ensure that the} \del{in order to make} DaPC NN training \rw{is consistent} \del{comparable} with the current practice for conventional DANN structures.
o ensure that the training of the DaPC NN is consistent with the current methodology used for conventional DANN structures. Thus, we consider the following optimization problem regarding the unknown weights forming the vector \rw{
$\bm{w}=\big\{w^{(\mathcal{L},\mathcal{N})}_{i},i=1,\ldots,N_w \big\}$ for all neurons distributed within the all layers}
: 
\begin{equation}
\label{OF}
\begin{aligned}
	 \bm{w} =\operatorname{arg\,min}_{\bm{w}} \biggl[&\frac{1}{T_N}\sum_{i=1}^{T_N} \left( \mathcal{R}(\bm{\omega_{T_i}})-R_{T_i} \right)^2\\ 
	 &+ \frac{1}{N_w}\sum_{i=1}^{N_w} \left(w^{(\mathcal{L},\mathcal{N})}_{i}\right)^2 \biggr],
\end{aligned}
\end{equation}
%
where the weights should be optimized \rw{simultaneously} to minimize the so-called loss function \cite{Goodfellow2016deeplearning} in equation (\ref{OF}). The first term in equation (\ref{OF}) reflects the deviation of the response $\mathcal{R}$ from the training data via the mean squared error (MSE). The second term corresponds to a regularization known as mean square weights (MSW) \footnote{See also Bayesian regularized \rw{neural networks} in \cite{okut2016bayesian})}. Other loss functions are also known in the literature, depending on the overall objective such as a Bayesian interpolation \cite{mackay1992bayesian}, physical regularization \cite{jia2021pgnn,praditia2020thermo} or classification tasks \cite{ballard1982computervision,krizhevsky2017cnn} \del{via cross-entropy between the prediction and the target distribution}. \rw{Moreover, the scalability to larger datasets relies on the chosen architecture and training methods. The effective model complexity proposed in \cite{nakkiran2021deep} identifies specific regimes where increasing the number of training samples may negatively impact performance.}

In order to directly compare the DaPC NN structure with the conventional DANN under identical conditions, we choose the Levenberg-Marquardt algorithm \cite{marquardt1963} to solve \rw{\footnote{We are aiming to not impose any preferences on the training algorithms in the current paper and we encourage the developers of training algorithms to incorporate their approaches to the DaPC NN framework.}} equation (\ref{OF}) for all structures due to its robustness. We also use the exact same training procedure for all structures using the exactly same loss-function in equation (\ref{OF}) and the exactly same training data sets $\bm{D}_{T_N}$. To provide gradients to the Levenberg-Marquardt algorithm, we use the analytical derivatives for DaPC NN introduced in \textit{Property 3} of Section \ref{DaPC_prop}, whereas we use typical backpropagation to provide gradients. \rw{Thus, the orthogonal bases in DaPC NNs are updated simultaneously with the weights during the training procedure (see also Figure \ref{fig:model}). Moreover, as pointed out in Section \ref{Deep_OPR}, the weights determine uniquely the corresponding data-driven orthonormal bases in equations (\ref{basis_multi}) and (\ref{basis_multi}) for each layer of DaPC NNs by solving the linear system of equations (\ref{orth_matrix_dd}) introduced in Section \ref{ML_aPC}. Consequently, the DaPC NN can be seen as a black box, where only the unknown weights need to be determined using one or another training procedure. This makes it easy for users to operate the DaPC NN in a similar way to the conventional DANN, without requiring them to have any special knowledge of its internal workings.}

The training procedure of ML approaches with single-layer ML representations \cite{deng2011an} like the PCE usually determines the unknown weights by solving a linear system of equations. Therefore, the unknown weights $w_{i}$ of the aPC representation in Section \ref{ML_aPC} are directly obtained by solving a linear system of equations (\ref{orth_matrix_dd}) (see details in \rw{\cite{Olad_al_CG2009,augustin2008pc,atkinson1992optimum}}). For numerical robustness, we use Moore-Penrose inversion \cite{moore1920reciprocal,penrose1956best, barata2012moore} (also known as pseudoinverse) to solve the regularized \cite{tikhonov1977solutions} least-squares problem of equation (\ref{OF}) as in the original aPC work \cite{OladNowak_RESS2012} that is available in Matlab \rw{file} exchange \cite{aPC_Toolbox}.

In the current paper, we \del{minimize speculations about the loss function and } focus on fundamental issues of deep neuronal network structures exploring the concept of orthonormal decomposition and the possibility to introduce high-order neural interaction, \rw{rather than delving into specifications of the loss function or training algorithm.}  Therefore, the reader is invited to adopt the introduced DaPC NN structure for own needs, introducing any own specific loss function or training procedure. The DaPC NN is available online for the reader through Matlab file exchange \cite{DaPC-NN_Toolbox}.

\subsubsection{Technical remark on DaPC NN bases}

Technically, the training procedure specified in equation (\ref{OF}) remains the same for the DaPC NN as well as for the DANN and the aPC, and the authors of the current paper do not aim at any novelty here. However, we would like to underline that the DaPC NN constructs its data-driven orthonormal bases implicitly during the training procedure, based on the input distribution and upon the responses of hidden nodes that react to adjustments of the weights during the training procedure. This means, the orthonormal bases on all layers are adaptively recomputed during the iterative training procedure while searching for the optimal weights. It could be seen as re-adjusting of orthonormal bases simultaneously within the iteration procedure over the weights. 

\section{Illustration of Performance and Comparison}
\label{Results}

In the previous Section \ref{ML}, we have introduced the DaPC NN using the data-driven orthonormal decomposition from aPC theory. The current section will illustrate the performance of the suggested DaPC NN, comparing it with the performance of a conventional DANN structure and also with an aPC expansion.
To do so, we will employ exactly the same training data set $\bm{D}_{T_N}=\big\{(\bm{\omega}_{Ti},R_{Ti}), i=1,\ldots,T_N \big\}$ of the size $T_N$ for all three ML approaches. The quality of the training itself (e.g. MSE) will not be strongly discussed as the training procedure provides only a small discrepancy between the ML model response $\mathcal{R}(\bm{\omega}_T)$ and the training data $\mathbf{R}_{T_N}$. \footnote{\del{We have observed for all considered test cases, that omitting regularization in loss function only slightly degrades the results of DaPC NN and aPC, whereas the conventional DANN shows extremely poor capability. Hence, pure comparison focusing on least-square solution without regularization seems to be not attractive.}} 
\rw{Such a discrepancy partially indicates whether the chosen architecture, loss function and training algorithm are appropriate for analysed problem.} Instead, we rather focus on the prediction ability for a separate data set that has not been used during the training procedure. Therefore, we will employ a validation data set \del{(also known as test data set)} \cite{james2014intro} $\bm{D}_{V_N}=\big\{(\bm{\omega}_{Vi},R_{Vi}), i=1,\ldots,V_N \big\}$ of the size $V_N$ with the validation inputs $\bm{\omega}_{V_N}=\{\bm{\omega}_{V_1},\ldots,\bm{\omega}_{V_N} \}$ and the corresponding validation responses $\mathbf{R}_{V_N}=\{R_{V_1},\ldots,R_{V_N}\}$.

Because the quality of the prediction is strongly dependent on the size $T_N$ of the available training data set \cite{OladNowak_RESS2012, hornik1989universal}, we will assess the prediction quality for validation data sets $\bm{D}_{V_N}$ under various sizes of the training data set $T_N$. To measure the prediction ability, we will assess the mean square error, which is usually used in ML, as well as a weighted error \rw{that will be defined below in eq.~\eqref{MSE},\eqref{ErrorMean},\eqref{ErrorStd}}, which is used in the uncertainty quantification community.  The mean square error could be also normalized by the variance, converting it to the coefficient of determination if desired. The corresponding estimates of the mean square error $MSE_{T_N}\left[\mathcal{R}(\bm{\omega}_{V_N})\right]$, the relative error of mean $E^\mu_{T_N}\left[\mathcal{R}(\bm{\omega}_{V_N})\right]$ and the relative error of standard deviation $E^\sigma_{T_N}\left[\mathcal{R}(\bm{\omega}_{V_N})\right]$ are defined as:
\begin{eqnarray}
\label{MSE}
	 \text{MSE}_{T_N}\left[\mathcal{R}(\bm{\omega}_{V_N})\right] = \frac{1}{V_N}\sum_{i=1}^{T_N} \left( \mathcal{R}(\bm{\omega_{V_i}})-R_{V_i} \right)^2,
\end{eqnarray}
\begin{eqnarray}
\label{ErrorMean}
E^\mu_{T_N}\left[\mathcal{R}(\bm{\omega}_{V_N})\right]= \frac{\mu\left[\mathcal{R}(\bm{\omega}_{V_N})\right]-\mu\left[\mathbf{R}_{V_N}\right]}{\mu\left[\mathbf{R}_{V_N}\right]}, 
\end{eqnarray}
\begin{eqnarray}
\label{ErrorStd}
E^\sigma_{T_N}\left[\mathcal{R}(\bm{\omega}_{V_N})\right] = \frac{\sigma\left[\mathcal{R}(\bm{\omega}_{V_N})\right]-\sigma\left[\mathbf{R}_{V_N}\right]}{\sigma\left[\mathbf{R}_{V_N}\right]}, 
\end{eqnarray}
where $\mu\left[\cdot\right]$ is the expected value and $\sigma\left[\cdot\right]$ is the standard deviation over the validation data set. The expected value $\mu[\cdot]$ and standard deviation $\sigma[\cdot]$ are obtained via 
the empirical mean and standard deviation over the validation set $\bm{D}_{V_N}$.

Because we are interested in the overall reliability of the ML response and to guarantee fairness in comparison, we omit the powerful property of aPC in computing the mean and variance without additional evaluations of the response. Instead, we will estimate them numerically using the final aPC response constructed on the validation data set, similarly to DANN and DaPC NN. 

As test cases, we use examples of functional approximation as in surrogate modeling. To test different aspects, we consider an Ishigami problem with three inputs \cite{Ishigami} in Section \ref{Ishigami}, an ON-10 problem with ten inputs \cite{oladyshkin2019connection} in Section  \ref{OladyshkinNowak} and also a carbon dioxide benchmark problem with non-linear shock propagation \cite{Koeppel2017} in Section \ref{CO2benchmark}. 

\rw{\subsection{Architecture and technical specification of ML models}} 
\del{Technical specifications of the considered ML models used in all test cases are presented in the Appendix A for the sake of completeness.} 
\rw{Architectures of the considered ML models used in the current and upcoming test cases are specified via the total number of layers $L$, corresponding numbers of nodes $N^{(\mathcal{L})}$ for each layer $\mathcal{L}$ as $[N^{(1)},\ldots,N^{(L)}]$, degrees of non-linearity $d^{(\mathcal{L})}$ for on each layer $\mathcal{L}$ as $[d^{(1)},\ldots,d^{(L)}]$,  activation functions $\mathcal{A}^{(\mathcal{L})}$ (same for all layers $\mathcal{L}$) and loss function ($\mathcal{LF}$). For the sake of transparency, we will also provide the total number of unknown weights or coefficients as $N_w$ in upcoming Sections. This is particularly useful for aPC and DaPC NN, as the number of unknown weights may not be as intuitive as for conventional DANN. We will adopt the architectures of the analyzed ML models taking into consideration the size $T_N$ of the training data sets. Nevertheless, we would like to emphasize that the setup of the corresponding architectures are the responsibility of the ML modeler. For our study, we selected the architectures with equal care and effort for all three compared approaches, trying to achieve the most reliable results. Nevertheless, the degrees of freedom in choosing the number of layers, the number of nodes per layer, the degree of non-linearity (if allowed) and the activation function pose strong challenges onto the modeling procedure, requiring a deep understanding of each underlying ML approach. For instance, in addition to the standard settings for the conventional DANN architecture, the DaPC NN architecture incorporates additional degrees of freedom by choosing the non-linearity degree on specific layers, which introduces unknown weights associated with the number of input neurons from the previous layer, as discussed in Section \ref{ML_DANN}. Exploring all possible configurations of deep architectures using a trial-and-error approach is not feasible due to the vast number of combinations. Consequently, developing an adaptive strategy for setting up the architecture could significantly improve the performance of deep multi-layer representations. This approach would require further research to minimize the potential for subjectivity in the modeling procedure. The authors of the current paper recognize the potential of the Bayesian framework for optimizing hyper-parameters such as the number of layers, number of nodes, and degree of non-linearity. However, direct Bayesian analysis seems to be computationally intensive, and it may be necessary to rely on approximate indicators that involve certain assumptions \cite{oladyshkin2019connection}. However, this topic goes beyond the scope of the current paper but could be the subject of future research.}

We encourage the reader to test the suggested DaPC NN approach and also known aPC and DANN approaches for own needs. The conventional DANN approach used here can be found under Matlab software \cite{MATLAB:2019b} using the \emph{fitnet} functionality. The aPC and DaPC NN approaches are available online for the reader through Matlab file exchange \cite{aPC_Toolbox} and \cite{DaPC-NN_Toolbox}, correspondingly.

\subsection{Ishigami test case}
\label{Ishigami}
\begin{table*} [!htb]
\caption{Architectures of aPC, DANN and DaPC NN for Ishigami test case.}
\label{IshigamiTestCase}       
\begin{tabular}{| c | c |c | c | c | c | c | c |}
    \hline
    ML model & $T_N$ & $L$ & $N^{(\mathcal{L})}$ & $d^{(\mathcal{L})}$ & $\mathcal{A}^{(\mathcal{L})}$ & $N_w$ & $\mathcal{LF}$\\
    \hline \hline 
     & 10 & 1 & 1 & 2 & none & 10 & MSE+MSW\\
     aPC & 100 & 1 & 1 & 4 & none & 35 & MSE+MSW\\
     & 1000 & 1 & 1 & 6 & none & 84 & MSE+MSW\\
     \hline 
     & 10 & 3 & [6,6,1] & [1,1,1] &  eq. (\ref{DANN_Tanh}) & 78 & MSE+MSW\\
     DANN & 100 & 4 & [9,6,3,1] & [1,1,1,1] & eq. (\ref{DANN_Tanh}) & 127 & MSE+MSW\\
     & 1000 & 4 & [10,8,6,1] & [1,1,1,1] & eq. (\ref{DANN_Tanh}) & 189 & MSE+MSW\\
     \hline
     & 10 & 2 & [3,1] & [2,2] & eq. (\ref{DaPC_normalization}) & 40 & MSE+MSW\\
     DaPC NN & 100 & 3 & [3,3,1] & [2,2,2] & eq. (\ref{DaPC_normalization}) & 70 & MSE+MSW\\
     & 1000 & 3 & [3,3,1] & [3,3,2] & eq. (\ref{DaPC_normalization}) & 130 & MSE+MSW\\
    \hline
\end{tabular}
\end{table*}

\subsubsection{Problem set up}
As the first test case, we will employ the widely used Ishigami function~\cite{Ishigami}, as it shows strong non-linearity accompanied with  non-monotonicity:
\begin{equation}
\mathcal{M}(\bm{\omega})=\sin \left(\omega_1\right)+a \sin ^{2}\left(\omega_2\right)+b \omega_3^{4} \sin \left(\omega_1\right),
\end{equation}
where we will use the particular case with $a = 7$ and $b = 0.1$. The distribution of the three input random variables $\bm{\omega}$ is given by mutually independent uniform distributions with $\omega_i\sim\mathcal{U}(-\pi,\pi)$. 
\begin{figure*}[!htb]
	\centering
	\includegraphics[width=0.66\linewidth]{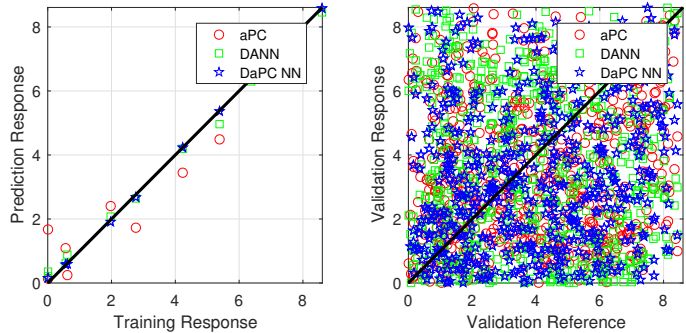}
	\caption{Prediction of aPC, DANN and DaPC NN for Training (left) and Validation (right) data: Ishigami test case with three inputs and the size of training data set equal to 10.}
 \label{Ishigami_3P_10T}
\end{figure*}
\begin{figure*}[!htb]
	\centering
	\includegraphics[width=0.66\linewidth]{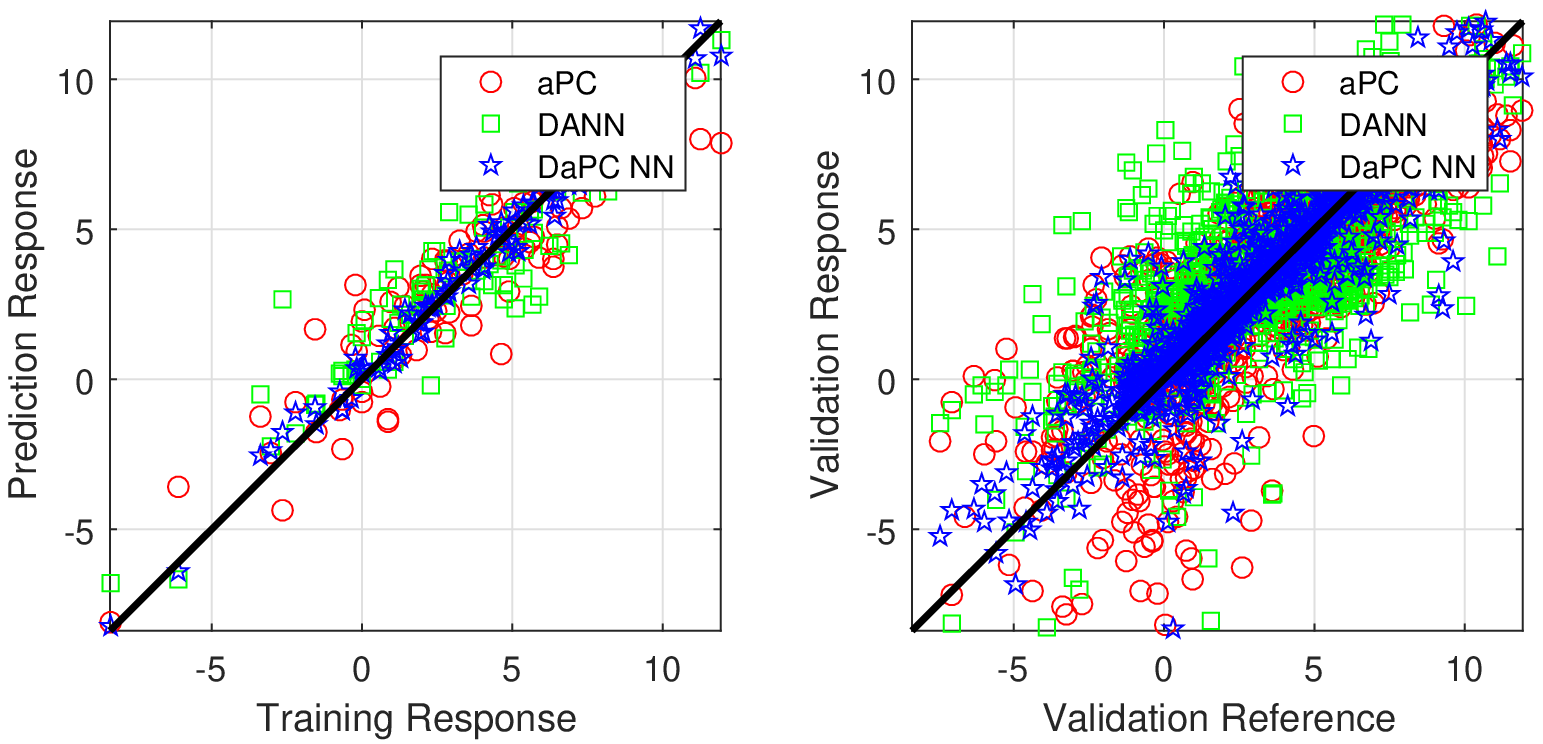}
	\caption{Prediction of aPC, DANN and DaPC NN for Training (left) and Validation (right) data: Ishigami test case with three inputs and the size of training data set equal to 100.}
 \label{Ishigami_3P_100T}
\end{figure*}
\begin{figure*}[!htb]
	\centering
	\includegraphics[width=0.66\linewidth]{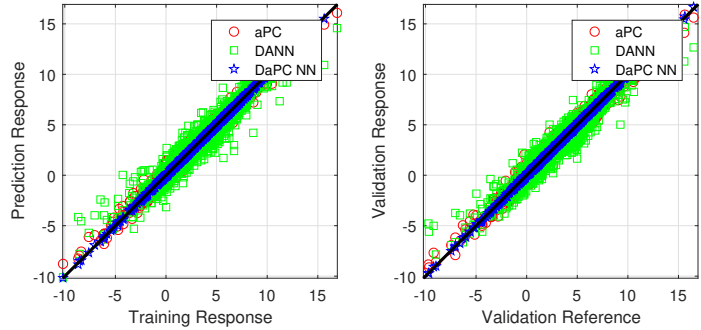}
	\caption{Prediction of aPC, DANN and DaPC NN for Training (left) and Validation (right) data: Ishigami test case with three inputs and the size of training data set equal to 1000.}
 \label{Ishigami_3P_1000T}
\end{figure*}

To obtain a training data set, we use Sobol sequences \cite{sobol2011construction} for the underlying distribution of inputs. To analyze how the quality of the prediction depends on the size of the training data set, we use the training data sets $\bm{D}_{T_N}$ of size $T_N$ equal to 10, 100 and 1000.  To asses\rw{s} the quality of prediction, we generate a validation data set $\bm{D}_{V_N}$ of the size $V_N=10^3$, via Monte Carlo sampling \cite{Smith1992}.

\subsubsection{Training and Validation}
Figures \ref{Ishigami_3P_10T}-\ref{Ishigami_3P_1000T} demonstrate how predictions made by aPC, conventional DANN and DaPC NN match the training and validation data sets for the size $T_N$ of training data equals to 10, 100 and 1000. In principle, all considered ML approaches show the necessary flexibility to \rw{approximate} \del{catch} the training data of different sizes, but the DaPC NN shows a superior performance with the smallest scatter in the left plots of Figures~\ref{Ishigami_3P_10T}-\ref{Ishigami_3P_1000T}. It is remarkable here that the DaPC NN is equipped with fewer degrees of freedom (number of weights) in comparison to the DANN \del{(see Table \ref{IshigamiTestCase})}. \rw{Table \ref{IshigamiTestCase} show the technical specifications for the Ishigami test case. Additionally, Table \ref{IshigamiTestCase} indicated the total number of unknown weights/coefficients by $N_w$ as for aPC and DaPC NN it could be less intuitive as for conventional DANN. Overall, the selection of the ML model architectures should take into account the size of the training data sets $T_N$ to provide certain flexibility during the training procedure.} 
\begin{figure*}[!htb]
	\centering
	\includegraphics[width=1.0\linewidth]{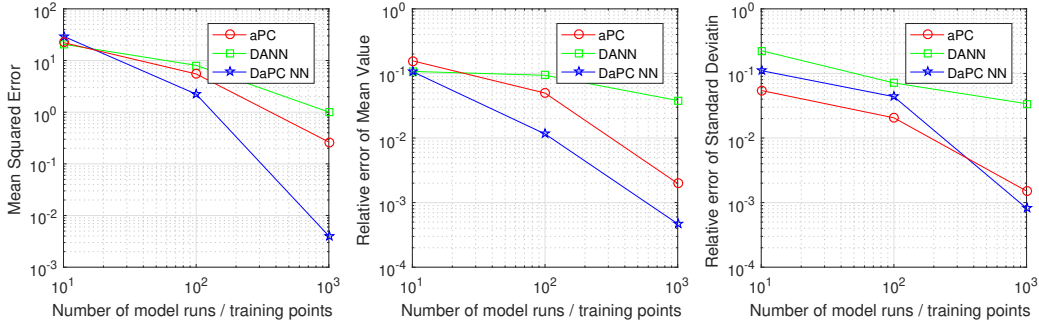}
	\caption{Performance of aPC, DANN and DaPC NN for Ishigami test case with three inputs: Convergence of Mean Square Error, Relative Error of Mean Value and Relative Error of Standard Deviation relatively the reference validation data set.}
 \label{Ishigami_3P_Convergence}
\end{figure*}

However, flexibility alone is not sufficient to make reliable prediction outside of the training data sample and \rw{only shows quality of the training approach} \del{hence, practically almost irrelevant}. Therefore, we pay stronger attention to the right plots in Figures~{\ref{Ishigami_3P_10T}-\ref{Ishigami_3P_1000T}}, reflecting the validation performance. All three approaches are powerless in case of the very small data set used for training (right plot in Figure \ref{Ishigami_3P_10T}). As expected, increasing the size of training data set to 100 (right plot in Figure~\ref{Ishigami_3P_100T}) or even 1000  (right plot in Figure~\ref{Ishigami_3P_1000T}) helps to overcome these issues during the validation phase. In particular, the DaPC NN demonstrates the best performance in the validation phase as well, even with a moderate sample size of training data.  Figures~\ref{Ishigami_3P_10T}-\ref{Ishigami_3P_1000T} indicate that the flexibility of DANN tends to have over-fitting issues regarding the regularisation as specified in equation (\ref{OF}). Opposite to theDANN, the DaPC NN shows less issues with over-fitting, inhering the orthonormality from the aPC representation and also the required flexibility from DANN. Due to its definition, the aPC itself seems to have not enough of flexibility to capture the validation data set as \del{good} \rw{well} as the DaPC NN. \rw{We also have observed in the current and upcoming test cases, that omitting regularization in loss function only slightly degrades the results of DaPC NN and aPC, whereas the conventional DANN shows extremely poor prediction. Hence, pure comparison focusing on least-square solution without regularization seems to be not attractive.}  

\subsubsection{Evidence of Convergence}

\rw{It Figures \ref{Ishigami_3P_10T}-\ref{Ishigami_3P_1000T}, it appears that DANNs have a stronger spreading of the point clouds and also higher density of spreading in comparison to DaPC NN and aPC models. However, determining the density of spreading based on visual inspection alone can be challenging. Therefore, we will pay closer attention the metrics introduced in the beginning of Section, which provides more clarity on this matter.}
Figure~\ref{Ishigami_3P_Convergence} shows the convergence in terms of mean square error, relative error of the mean value and relative error of the standard deviation as a function of the training data size $T_N$, applied to the validation data set $\bm{D}_{V_N}$ for aPC expansion, conventional DANN and DaPC NN. The figure reveals that a faster convergence has been reached for DaPC NN in terms of all investigated convergence metrics, summarising the observations made in the previous section. The relative errors of mean and the standard deviation are smaller with the aPC than with the conventional DANN.

\subsection{ON-10 test case}
\label{OladyshkinNowak}
\begin{table*}[!htb]
\caption{Architectures of aPC, DANN and DaPC NN for ON-10 test case.}
\label{ONTestCase}       
\begin{tabular}{| c | c |c | c | c | c | c | c |}
    \hline
    ML model & $T_N$ & $L$ & $N^{(\mathcal{L})}$ & $d^{(\mathcal{L})}$ & $\mathcal{A}^{(\mathcal{L})}$ & $N_w$ & $\mathcal{LF}$\\
    \hline \hline 
    &100 & 1 & 1 & 2 & none & 66 & MSE+MSW\\
    aPC &500 & 1 & 1 & 3 & none & 286 & MSE+MSW\\
    &1000  & 1 & 1 & 4 & none & 1001 & MSE+MSW\\
    \hline
    & 100 & 4 & [9,6,3,1] & [1,1,1,1] & eq. (\ref{DANN_Tanh}) & 175 & MSE+MSW\\
    DANN & 500 & 4 & [15,10,5,1] & [1,1,1,1] & eq. (\ref{DANN_Tanh}) & 386 & MSE+MSW\\
    & 1000 & 4 & [15,10,5,1] & [1,1,1,1] & eq. (\ref{DANN_Tanh}) & 386 & MSE+MSW\\
    \hline
    & 100 & 2 & [5,1] & [2,3] & eq. (\ref{DaPC_normalization}) & 386 & MSE+MSW\\
    DaPC NN & 500 & 2 & [10,1] & [2,3] & eq. (\ref{DaPC_normalization}) & 946 & MSE+MSW\\
    & 1000 & 2 & [10,1] & [2,3]  & eq. (\ref{DaPC_normalization}) & 946  & MSE+MSW\\
    \hline
\end{tabular}
\end{table*}

\subsubsection{Problem set up}
\label{SetUpON}

As second test case, we will consider a non-linear analytical function $\mathcal{M}(\bm{\omega})$ of ten ($n=10$) uncertain inputs $\bm{\omega}=\left\{\omega_1,\ldots,\omega_n \right\}$ from the paper \cite{oladyshkin2019connection}:
\begin{equation}
\label{problemON}
\begin{aligned}
\mathcal{M}(\bm{\omega})=&(\omega_1^2+\omega_2-1)^2+\omega_1^2+0.1\omega_1\exp(\omega_2)\\
&+1+\sum_{i=3}^{n}\frac{\omega_i^3}{i},
\end{aligned}
\end{equation}
where the inputs $\bm{\omega}$ in equation (\ref{problemON}) are considered to be independent uniformly distributed with $\omega_i\sim\mathcal{U}(-5,5)$ for $i=1 \ldots 10$. 

For the current test case, we will consider training data sets $\bm{D}_{T_N}$ of the sizes $T_N$ equal to $100$, $500$ and $1000$. Similar to the previous test case, we will generate our training data according to the Sobol sequence \cite{sobol2011construction} based on the underlying distributions of inputs $\bm{\omega}$. As before, we employ a validation data set $\left\{(\bm{\omega}_{Vi},R_{Vi}), i=1,\ldots,V_N \right\}$ of the size $V_N=10^3$ generated by Monte Carlo sampling \cite{Smith1992} from the distribution of inputs $\bm{\omega}$.

\subsubsection{Prediction and Validation}
\label{Pred_ON}

\begin{figure*}[!htb]
	\centering
	\includegraphics[width=0.66\linewidth]{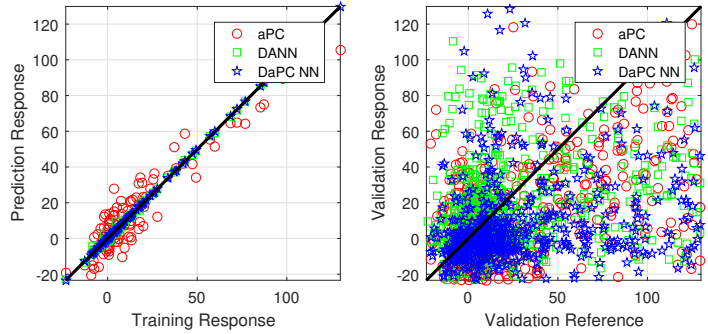}
	\caption{Prediction of aPC, DANN and DaPC NN for Training (left) and Validation (right) data: ON-10 test case with ten inputs and the size of training data set equal to 100.}
 \label{TestProblem_10P_100T}
\end{figure*}

\begin{figure*}[!htb]
	\centering
	\includegraphics[width=0.66\linewidth]{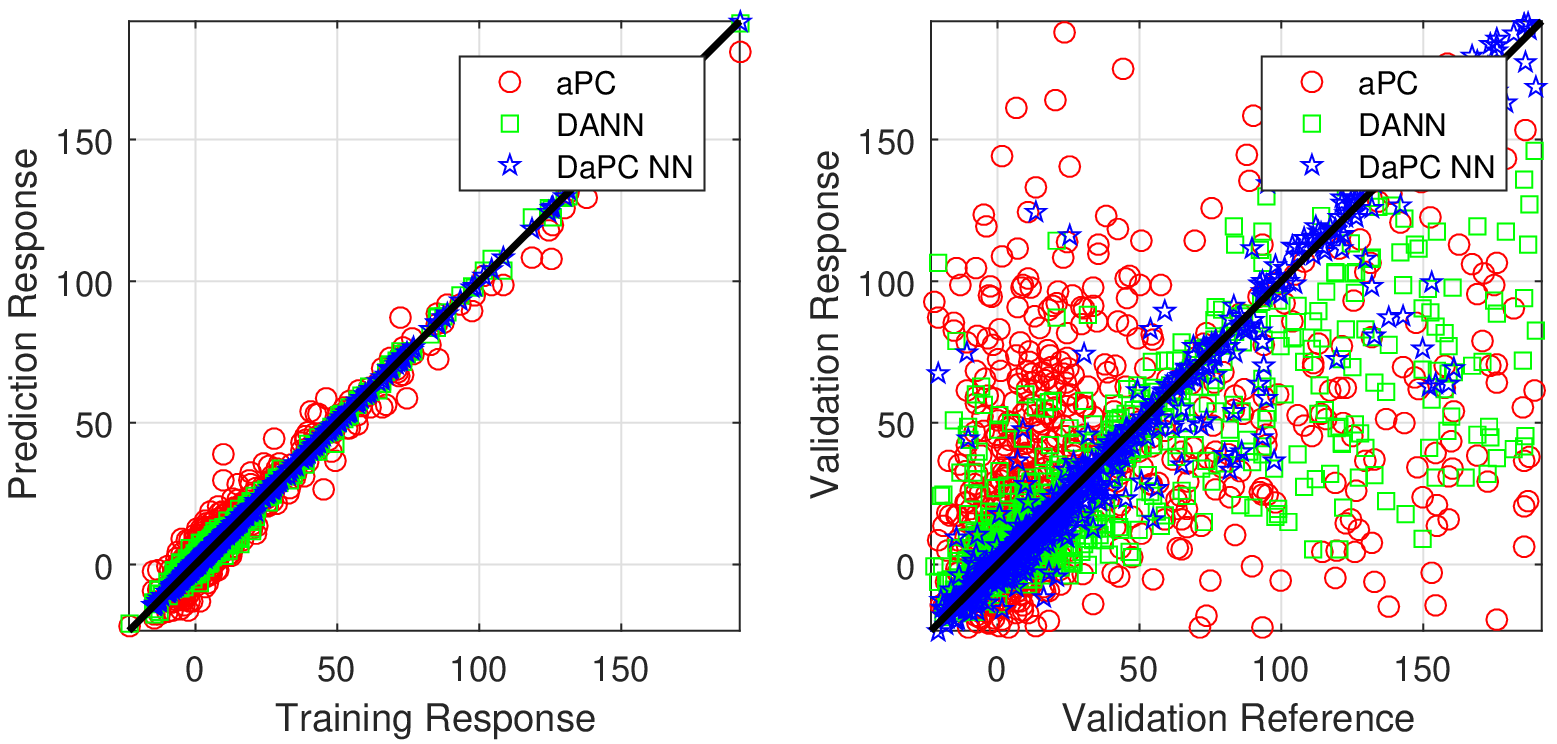}
	\caption{Prediction of aPC, DANN and DaPC NN for Training (left) and Validation (right) data: ON-10 test case with ten inputs and the size of training data set equal to 500.}
 \label{TestProblem_10P_500T}
\end{figure*}

\begin{figure*}[!htb]
	\centering
	\includegraphics[width=0.66\linewidth]{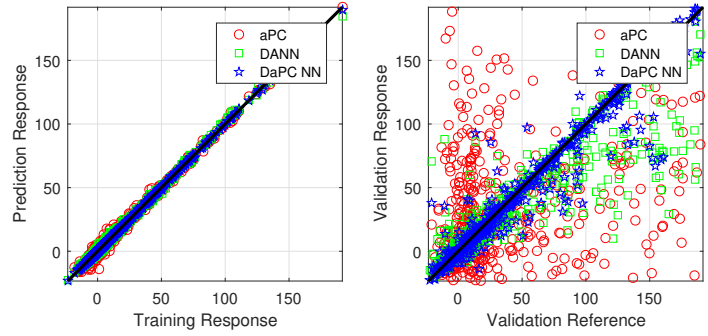}
	\caption{Prediction of aPC, DANN and DaPC NN for Training (left) and Validation (right) data: ON-10 test case with ten inputs and the size of training data set equal to 1000}
 \label{TestProblem_10P_1000T}
\end{figure*}

Figures \ref{TestProblem_10P_100T}-\ref{TestProblem_10P_1000T} show the predictions made by the aPC, DANN, and DaPC NN compared with the corresponding data during the training and validation phases. Architectures of the considered ML models are presented in Table \ref{ONTestCase} of Appendix A. With a low number of training data (Figure \ref{TestProblem_10P_100T}), the DANN and DaPC NN a show better ability to capture the training data compared to the aPC expansion. Here, the aPC demonstrates a lack of flexibility during training. During validation, however, all methods fail to produce consistent results when the small set is used as shown in Figure \ref{TestProblem_10P_100T}. When trained with more data (Figure \ref{TestProblem_10P_500T} and Figure \ref{TestProblem_10P_1000T}), DaPC NN produces predictions with relatively high accuracy compared to aPC and DANN even during validation; where aPC and DANN do not benefit significantly from the increase of training data. Moreover, the orthonormal structure of the DaPC NN representation handles the overfitting issues extremely well. \rw{Table \ref{ONTestCase} presents the technical specifications for the ON-10 test case, including the total number of unknown weights/coefficients denoted by $N_w$ for all ML models. Again, when selecting an appropriate ML model architecture, it is important to consider the size of the training data set $T_N$ that can influence the flexibility of the training process and the predictive power of the model.} For example, the number of unknown weights in the DaPC NN \del{could} \rw{can} be two times higher than the size of the training data.

It seems that the high non-linearity of the considered test case strongly limits the aPC to obtain the necessary polynomial representation. A high degree of freedom is very helpful\del{ly} in this case. Opposite to that, the DANN seems to suffer significantly from over-fitting, regardless of the regularization in that scenario. Overall, the DaPC NN predictions are more consistent, with lower scatter compared to the other methods for the considered test case.

\subsubsection{Evidence of convergence}
\label{Ev_ON}
\begin{figure*}[!htb]
	\centering
	\includegraphics[width=1.0\linewidth]{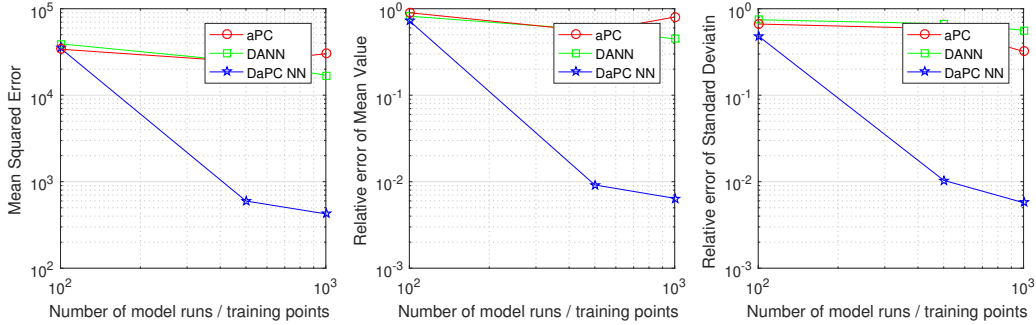}
	\caption{Performance of aPC, DANN and DaPC NN for ON-10 test case with ten inputs: Convergence of Mean Square Error, Relative Error of Mean Value and Relative Error of Standard Deviation relatively the reference validation data set.}
 \label{TestProblem_10P_Convergence}
\end{figure*}

Figure \ref{TestProblem_10P_Convergence} shows convergence in \del{therms} \rw{terms} of mean square error, relative error of mean value and relative error of standard deviation over the training data size $T_N$. Both the aPC and DANN converge to a similar point, meaning that the performance of both methods \del{are} \rw{is} comparable. Figure \ref{TestProblem_10P_Convergence} also indicates that increasing the number of training points does not improve the aPC and DANN performance significantly, again confirming our observation in the previous Section \ref{Pred_ON}. The DaPCNN, on the other hand, distinctly outperforms the aPC and DANN, with approximately two orders of magnitudes lower prediction errors in validation. The effect of increasing the number of training points can be seen clearly in the DaPCNN performance, showing better learning capacity compared to aPC and DANN.

\subsection{CO$_2$ Benchmark problem}
\label{CO2benchmark}
\begin{figure*}[!htb]
	\centering
	\includegraphics[width=0.8\linewidth]{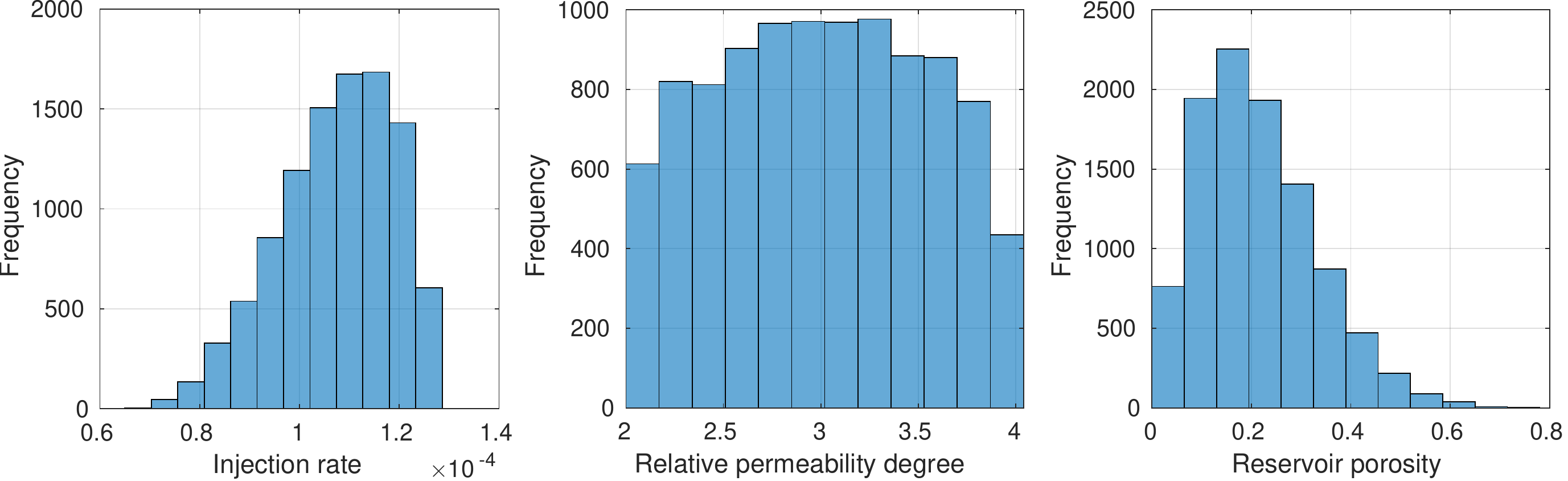}
	\caption{Distributions of injection rate ($m^3/s$), relative permeability degree (-), and reservoir porosity (-)}
 \label{co2_params}
\end{figure*}

\subsubsection{Benchmark set up}
As third test case, we will consider a carbon dioxide (CO$_2$) benchmark model that has already been used to compare different ML approaches in the earlier paper \cite{Koeppel2017}. The particularity of that problem consists in a strong shock propagation, where various ML approaches relying on smooth functions could suffer a lot. The problem refers to a multi-phase flow in porous media, where  CO$_2$ is injected into a deep aquifer and then spreads in a geological formation. This yields a pressure build-up and a plume evolution. CO$_2$ injection into the subsurface could be a possible practice to mitigate the CO$_2$ emission into the atmosphere. The CO$_2$ benchmark model proposed by Köppel et al. \cite{koppel2019comparison} is a reduced version of the model in a benchmark problem defined in the paper \cite{class2009benchmark}. This reduction consists of a radial flow in the vicinity of the injection well, and \rw{is} made primarily due to the high computational demand of the original CO$_2$ model.  It is assumed that the fluid properties such as the density and the viscosity are constant, and all processes are isothermal. The CO$_2$ and the brine build two separate and immiscible phases, and mutual dissolution is neglected. Additionally, the formation is isotropically rigid and chemically inert, and capillary pressure is negligible. We consider the CO$_2$ saturation to be the quantity of interest as a model response that is a function of the coordinates for space $x$ and time $t$ as introduced in \cite{koppel2019comparison}. Overall, the considered CO$_2$ benchmark problem is strongly non-linear because the CO$_2$ saturation spreads as a strongly non-linear front that could be challenging to capture via surrogates. For detailed information on the governing equations, the modeling assumptions and numerical approaches, the reader is referred to the original publication \cite{koppel2019comparison}. 

Following the comparison study \cite{koppel2019comparison}, we consider the combined effects of three sources of uncertainty.  We take into account the uncertainty of boundary conditions via the injection rate, the uncertainty of constitutive relation introduced via in the relative permeability definitions and the uncertainty of material properties represented by the porosity of the geological formation. Figure \ref{co2_params} shows the distribution of these three inputs taken from the original data set \cite{BenchmarkData}. 

\begin{table*}
\caption{Architectures of aPC, DANN and DaPC NN for CO$_2$ benchmark problem employing Sobol sequences.}
\label{CO2TestCase}       
\begin{tabular}{| c | c |c | c | c | c | c | c |}
    \hline
    ML model & $T_N$ & $L$ & $N^{(\mathcal{L})}$ & $d^{(\mathcal{L})}$ & $\mathcal{A}^{(\mathcal{L})}$ & $N_w$ & $\mathcal{LF}$\\
    \hline \hline 
    & 100 & 1 & 1 & 3 & none & 20 & MSE+MSW\\
    aPC & 500 & 1 & 1 & 5 & none & 56 & MSE+MSW\\
    & 1000 & 1 & 1 & 10 & none & 286 & MSE+MSW\\
    \hline
    & 100 & 3 & [8,6,1] & [1,1,1,1]  & eq. (\ref{DANN_Tanh}) & 93 & MSE+MSW\\
    DANN & 500 & 3 & [8,6,1] & [1,1,1,1] & eq. (\ref{DANN_Tanh}) & 93 & MSE+MSW\\
    & 1000 & 4 & [10,8,6,1] & [1,1,1,1] & eq. (\ref{DANN_Tanh}) & 189 & MSE+MSW\\
    \hline  
    & 100 & 2 & [3,1] & [2,2] & eq. (\ref{DaPC_normalization}) & 40 & MSE+MSW\\
    DaPC NN & 500 & 2 & [3,3] & [2,2] & eq. (\ref{DaPC_normalization}) & 80 & MSE+MSW\\
    & 1000 & 2 & [3,1] & [3,3] & eq. (\ref{DaPC_normalization}) & 80 & MSE+MSW\\
    \hline      
\end{tabular}
\end{table*}
Similar to the previous test cases, we will train the aPC, DANN and DaPC NN employing the exactly same training data sets $\bm{D}_{T_N}$ while varying the size $T_N$.  However, the final trained ML representation could depend not only on the size $T_N$ of the training data, but also on how the training data have been constructed. Indeed, at least for the polynomial representation, the training points must be selected with dedicated strategies in order to avoid additional oscillations known as the Runge phenomenon \cite{runge1901empirische}. In particular, wrong training points could lead to a very strong oscillations near the discontinuities (Gibbs effect) of the considered CO$_2$ benchmark. Therefore, it will be also interesting to see how the Runge phenomenon may affect the quality of DANN and DaPC NN training. 

In order to address that question, we will consider two training strategies.  In the first training strategy, we will use the Sobol sequence \cite{sobol2011construction} to construct the training sets of sizes $T_N$ equals to 100, 500 and 1000 similarly to the previous test cases. In the second training strategy, we will employ Gaussian integration points \cite{villadsen1978solution,MR2061539} to generate the training data sets of sizes $T_N$ equal to 27, 216 and 1331 according to the available distribution \cite{OladNowak_RESS2012} of inputs displayed in Figure \ref{co2_params}. In short, the optimal choice \cite{villadsen1978solution} of training points \rw{(i.e. Gaussian quadrature rule)} corresponds to the roots of the polynomial of one degree higher than the order used in the polynomial representation. The Gaussian integration points form a full tensor (FT) grid in the space of inputs. This is what leads to sizes of the training data sets equal to 27, 216 and 1331 related to the polynomial representation of 3$^{rd}$, 5$^{th}$ and 10$^{th}$ degrees correspondingly. Strictly, this training rule can be fully satisfied for aPC representation \rw{following the Gaussian quadrature rule \cite{villadsen1978solution}}, but there is no proof for multi-layered structures such as DANN or DaPC NN indicating that the corresponding Gaussian training strategy could mitigate the Runge phenomenon. To assess the quality of prediction, we will employ the validation data set $\bm{D}_{V_N}$ of the size $V_N=10^4$ \cite{BenchmarkData} generated according to the variability of the inputs (Monte Carlo approach) shown in Figure \ref{co2_params}.

\subsubsection{Prediction and Validation}

\begin{table*}
\caption{Architectures of aPC, DANN and DaPC NN for CO$_2$ benchmark problem employing Gaussian integration points.}
\label{CO2TestCaseG}       
\begin{tabular}{| c | c |c | c | c | c | c | c |}
    \hline
    ML model & $T_N$ & $L$ & $N^{(\mathcal{L})}$ & $d^{(\mathcal{L})}$ & $\mathcal{A}^{(\mathcal{L})}$ & $N_w$ & $\mathcal{LF}$\\
    \hline \hline 
    &27 & 1 & 1 & 2 & none & 10 & MSE+MSW\\
    aPC &216 & 1 & 1 & 5 & none & 56 & MSE+MSW\\
    &1331 & 1 & 1 & 10 & none & 286 & MSE+MSW\\
    \hline
    & 27 & 3 & [8,6,1] & [1,1,1,1] & eq. (\ref{DANN_Tanh}) & 93 & MSE+MSW\\
    DANN & 216 & 3 & [8,6,1] & [1,1,1,1] & eq. (\ref{DANN_Tanh}) & 93 & MSE+MSW\\
    & 1331 & 4 & [10,8,6,1] & [1,1,1,1] & eq. (\ref{DANN_Tanh}) & 189 & MSE+MSW\\
   \hline
    & 27 & 2 & [3,1] & [2,2] & eq. (\ref{DaPC_normalization}) & 40 & MSE+MSW\\
    DaPC NN & 216 & 2 & [3,1] & [2,2] & eq. (\ref{DaPC_normalization}) & 40 & MSE+MSW\\
    & 1331 & 2 & [3,1] & [3,3] & eq. (\ref{DaPC_normalization}) & 80 & MSE+MSW\\
    \hline
\end{tabular}
\end{table*}
We will reproduce the CO$_2$ saturation along the radial distance from the injection well for the fixed time instance of 100 days in accordance with the CO$_2$ benchmark scenario \cite{koppel2019comparison}. To do so, we will train aPC expansion, conventional DANN and DaPC NN for each numerical discretion cell (i.e. 250 times). In this sense, we will construct 250 networks/expansions that seek to capture features of CO$_2$ displacement in the subsurface. The related ML architectures are presented in Table \ref{CO2TestCase} \del{of Appendix A}. Figure \ref{CO2_Benchmark_Sobol_1000T} demonstrates the performance of the considered ML models during validation, using the Sobol sequence training data with $T_N=1000$. All three approaches are far away in capturing the reference values of mean and standard deviation for CO$_2$ saturation even considering a sufficiently large size of the training data. All three ML models suffer strongly from the Gibbs effect \rw{(i.e oscillations)} caused by the strong non-linearity of shock propagation (see examples in \cite{koppel2019comparison}). The regularization in equation (\ref{OF}) seems to show the most significant effect for the aPC. Opposite to that, using Gaussian integration points as training data helps to mitigate the Runge phenomenon and to reduce the Gibbs effect substantially not only for the aPC as expected, but also for the DaPC NN. Corresponding numbers of training points and technical specifications of the considered ML models are presented in Table \ref{CO2TestCaseG} \del{of Appendix A}. Figures \ref{CO2_Benchmark_FT_27T}-\ref{CO2_Benchmark_FT_1331T} demonstrate the predictions made by aPC expansion, conventional DANN and DaPC NN. It seems that the DaPC NN inherits partially the aPC properties that help overcome the Runge phenomenon, and the optimal training strategy relying on the distribution of the inputs helps to construct acceptable ML model at low computational costs (e.g. Figures \ref{CO2_Benchmark_FT_27T} or Figures \ref{CO2_Benchmark_FT_216T}).   Unfortunately, the training strategy based on Gaussian integration points does not improve the performance of the conventional DANN.  Indeed, as we have clarified in Section \ref{DANNmonomials}, the conventional DANN structure relies on Gaussian distribution of inputs with zero mean and unit variance, which is not fulfilled in the considered CO$_2$ benchmark case.  
\begin{figure*}[!htb]
	\centering
	\includegraphics[width=0.66\linewidth]{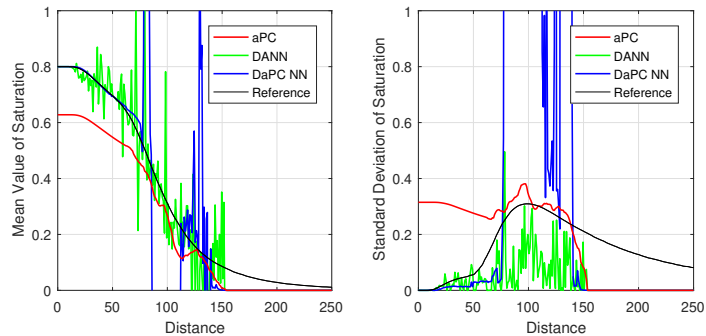}
	\caption{Prediction of mean (left) and standard deviation (right) using aPC, DANN and DaPC NN for Validation data: CO$_2$ benchmark problem using 1000 Sobol sequences as training data set.}
 \label{CO2_Benchmark_Sobol_1000T}
\end{figure*}
\begin{figure*}[!htb]
	\centering
	\includegraphics[width=0.66\linewidth]{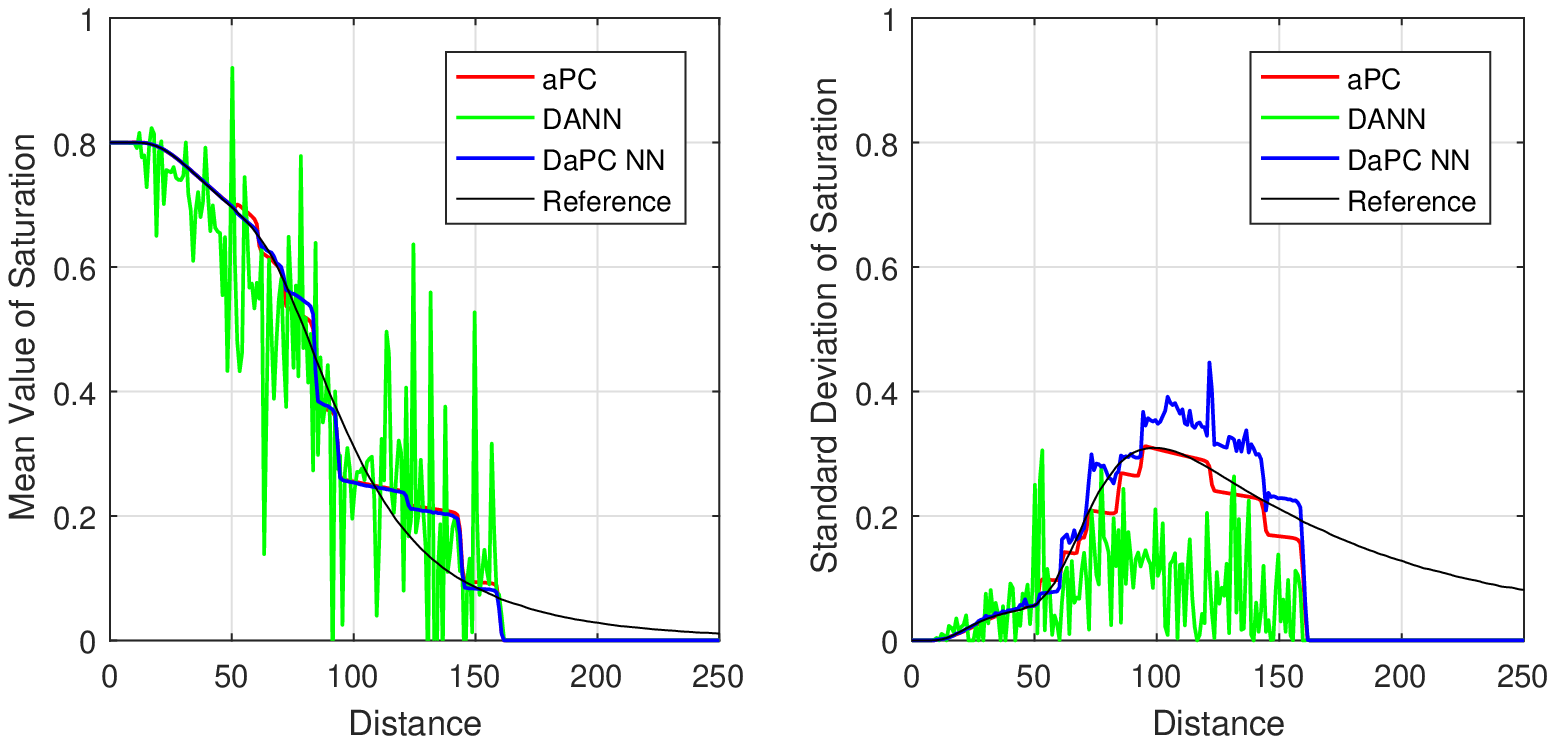}
	\caption{
	Prediction of mean (left) and standard deviation (right) using aPC, DANN and DaPC NN for Validation data: CO$_2$ benchmark problem using 27 Gaussian integration points as training data set.}
 \label{CO2_Benchmark_FT_27T}
\end{figure*}
\begin{figure*}[!htb]
	\centering
	\includegraphics[width=0.66\linewidth]{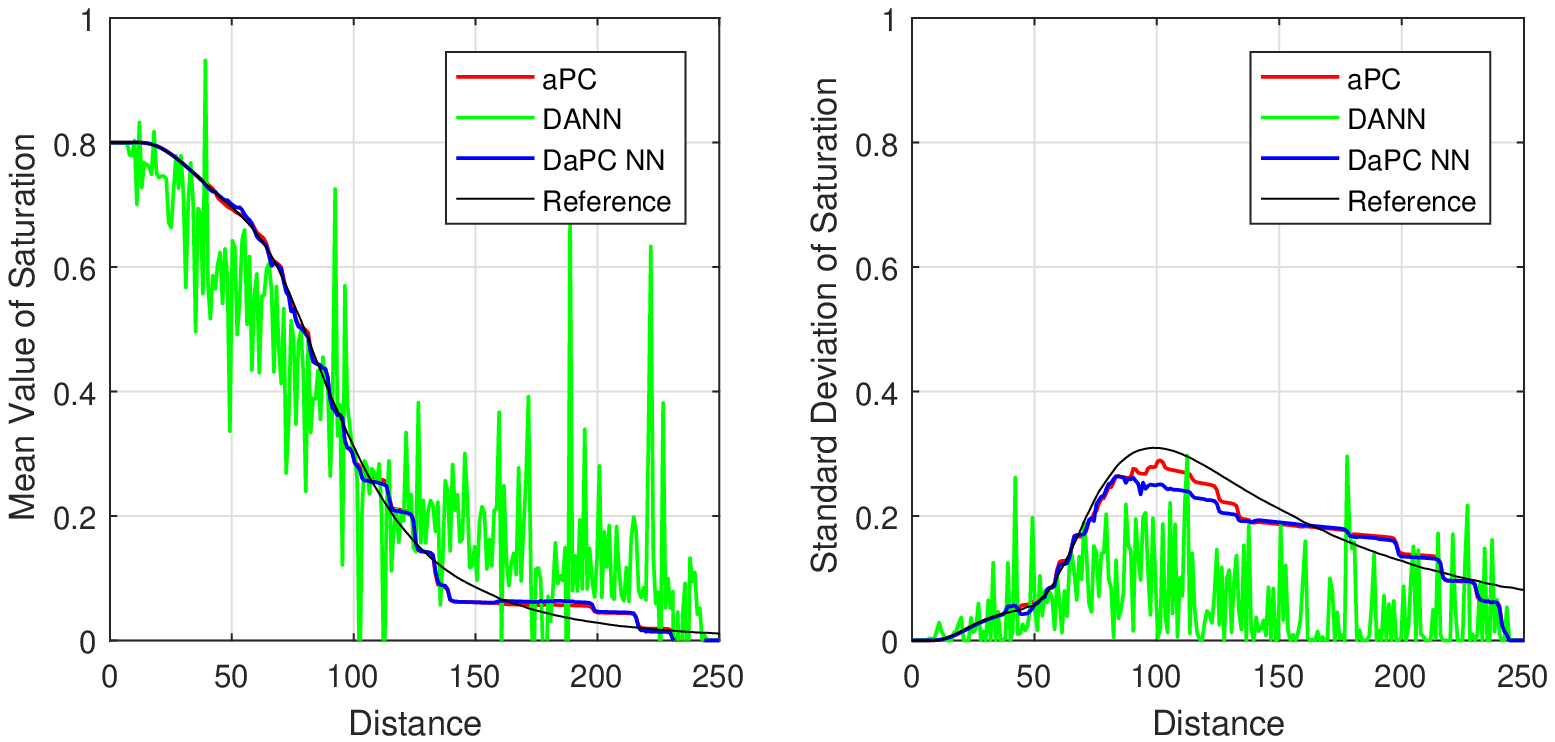}
	\caption{Prediction of mean (left) and standard deviation (right) using aPC, DANN and DaPC NN for Validation data: CO$_2$ benchmark problem using 216 Gaussian integration points as training data set.}
 \label{CO2_Benchmark_FT_216T}
\end{figure*}
\begin{figure*}[!htb]
	\centering
	\includegraphics[width=0.66\linewidth]{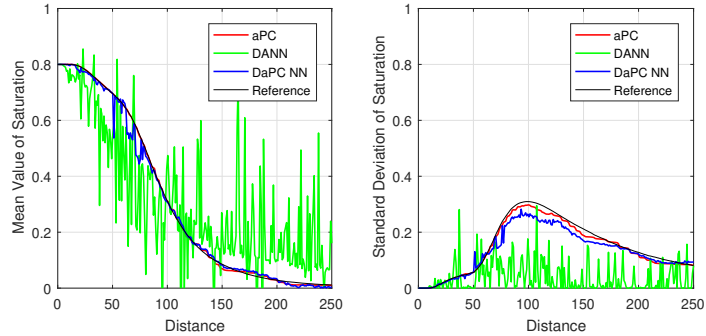}
	\caption{Prediction of mean (left) and standard deviation (right) using aPC, DANN and DaPC NN for Validation data: CO$_2$ benchmark problem using 1331 Gaussian integration points as training data set.}
 \label{CO2_Benchmark_FT_1331T}
\end{figure*}
\subsubsection{Evidence of convergence}
We will assess the convergence in terms of mean square error, mean value and standard deviation of CO$_2$ saturation over the training data size $T_N$ for both previously mentioned training strategies. In order to make the quantities of interest in correspondence with the original benchmark study \cite{Koeppel2017} and its follow-up \cite{rehme2021b}, we will consider the following space-averaged quantities:
\begin{equation}
\label{MSE_E}
    \text{MSE}_{T_N}= \frac{1}{250}\cdot\frac{1}{V_N} \sum_{i=1}^{V_N} \left\|\mathcal{R}(\bm{\omega}_{V_i}) - R_{V_i} \right\|_{L^2}^2,
\end{equation}
\begin{eqnarray}
\label{ErrorMean_E}
E^\mu_{T_N}= \frac{1}{250}\left\|\mu[\mathcal{R}(\bm{\omega}_{V_N})] - \mu[{\mathbf{R}_{V_N}}] \right\|_{L^2},
\end{eqnarray}
\begin{eqnarray}
\label{ErrorStd_E}
E^\sigma_{T_N}= \frac{1}{250}\left\|\sigma[\mathcal{R}(\bm{\omega}_{V_N})] - \sigma[\mathbf{R}_{V_N}] \right\|_{L^2}.
\end{eqnarray}

\begin{figure*}[!htb]
	\centering
	\includegraphics[trim=1.5cm 0cm 2.5cm 0cm, clip=true, width=0.9\linewidth]{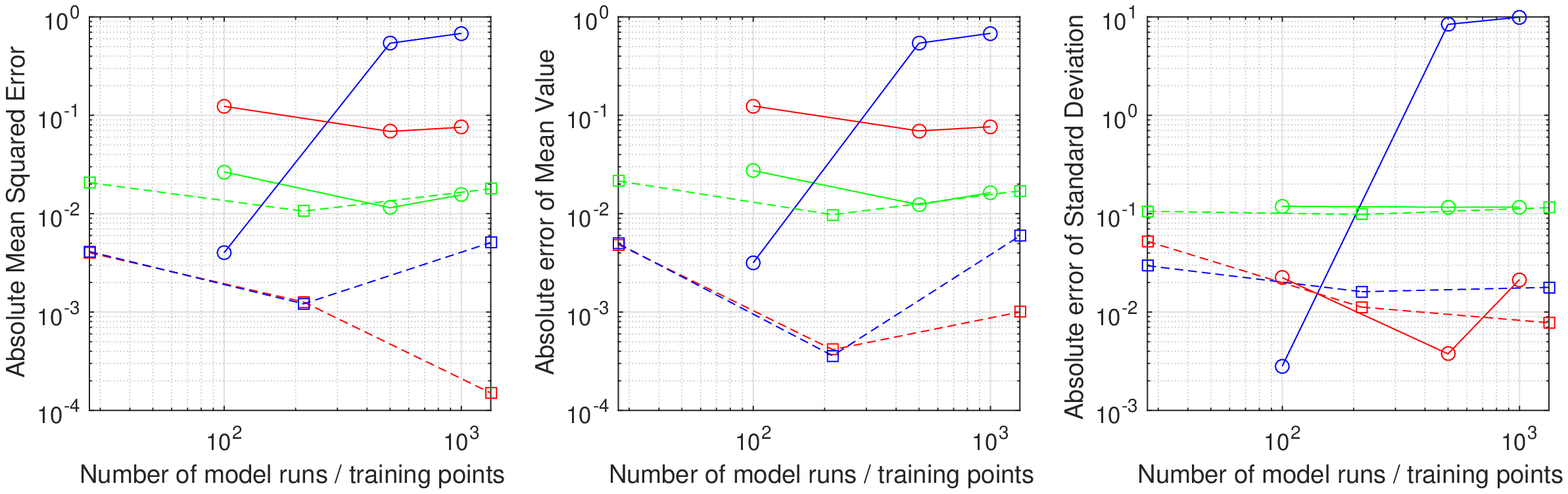}
    \includegraphics[trim=0cm 0cm 1.4cm 0cm, clip=true, width=0.09\linewidth]{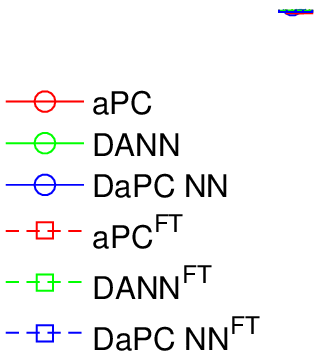}
	\caption{Performance of aPC, DANN and DaPC NN trained on Sobol sequences and Gaussian integration points for CO$_2$ benchmark problem: Convergence of Absolute Averaged Mean Square Error, Absolute Averaged Error of Mean Value and Absolute Averaged Error of Standard Deviation relatively the reference validation data set.}
 \label{CO2_Benchmark_Convergence}
\end{figure*}

Figure \ref{CO2_Benchmark_Convergence} shows convergence in terms of the absolute averaged mean square error in equation (\ref{MSE_E}), absolute averaged error of mean value in equation (\ref{ErrorMean_E}) and absolute averaged error of standard deviation in equation (\ref{ErrorStd_E}) over the training data size $T_N$ for aPC, conventional DANN and DaPC NN. Figure \ref{CO2_Benchmark_Convergence} displays the convergence of the analyzed ML approaches trained on the Sobol sequences and as well on the Gaussian integration points (marked by superscript FT).

We observe that the performance of DANN does not profit from an increasing training data size. Also, it does not benefit from the considered choices of training points and seems to be affected by oscillations caused by the strong non-linearity of the underlying problem. Additionally, increasing the training data size does not help the aPC expansion if the training points are not optimally distributed (see more details in \cite{OladNowak_RESS2012}), but it is clearly visible how the use of Gaussian integration points help overcome the Runge phenomenon. Moreover, the Gaussian integration points strongly help in DaPC NN training and also reduce the Gibbs effect, rendering the results more reliable. We also would like to remark, that turning the DaPC NN architecture into a one-layer structure reproduces the aPC results under the same technical specification. However, we have selected the multi-layered DaPC NN architecture as specified in Table \ref{CO2TestCaseG} for illustration of performance in order to provide the broader picture. Overall, we conclude that all considered ML approaches suffer a lot in capturing the strong shock propagation in the CO$_2$ benchmark model, as all three rely on smooth functions.

\subsection{Final Remarks}
In the current paper, we offer a view on neural signal processing in deep artificial neural networks from the PCE perspective, introducing in that way the DaPC NN. By employing PCE fundamentals, we demonstrate that orthonormality conditions in each node of the conventional DANN structure are fulfilled if the neural signal propagating through the multi-layer architecture was normally distributed with zero mean and unit variance. This situation is not necessarily satisfied for the majority of data-driven applications and, hence, could lead to redundant representation, where one neural signal could contain partial information that is also coming from other neurons. Moreover, introducing the PCE into the DANN structure provides an opportunity to go beyond the linear weighted superposition of single \rw{univariate} neurons on each node.

From the modelling perspective, the user is prompted to specify the DaPC NN architecture through the number of layers, number of nodes per layer, activation function and loss function similar as in a conventional DANNs. Additionally, the novel DaPC NN requires specification of the desired degree of non-linearity for each hidden layer. The latter aspect leads to high-order weighted superposition on each node of the network, so that non-linear activation functions become optional. This reduces the potential for subjectivity in the modeling procedure. \rw{However, it also introduces a new responsibility of choosing the degree of non-linearity, which in turn raises new research questions.}

Technically, the DaPC NN requires a similar training procedure as any conventional DANN, and all trained weights determine automatically the corresponding multi-variate data-driven orthonormal bases for all layers of DaPC NN. The DaPC NN Matlab Toolbox is available online and users are invited to adopt it for own needs \del{[CITE]} \rw{\cite{DaPC-NN_Toolbox}}.

To illustrate the performance of the DaPC NN, we have \del{conceded} \rw{investigated} three test cases, comparing the DaPC NN with the conventional DANN and also with aPC expansion. 
To do so, we have employed identical training procedures for DANN, DaPC NN and aPC, minimizing the mean squared error, regularized via ridge regression. In the training procedure, we employ exactly the same training and validation data sets for all three ML approaches. Additionally, we vary the size of the training data set and assess the convergence of the different ML approaches on a validation data set by evaluating the mean square error, error of mean and error of standard deviation.

In the first and second test cases, we observe that the DaPC NN reaches superior results in comparison to the aPC and DANN, even considering that we use only a moderate training data size. In the third CO$_2$ benchmark test case, we observe that all considered ML approaches suffer a lot in capturing the strong CO$_2$ shock propagation, as they all rely on smooth functions. However, using Gaussian integration points as training dataset helps to mitigate the Runge phenomenon and reduce the Gibbs effect substantially \del{not-only} \rw{not only} for the aPC but also for the DaPC NN. This emphasizes that the DaPC NN inherits partially the aPC properties that help to overcome the Runge phenomenon. Therefore, an optimal training strategy relying on the distribution of the inputs helps to construct an acceptable ML model at low computational costs.  Unfortunately, the performance of the DANN does not profit from the considered choices of training points. Increasing the training data size does not seem to be helpful against oscillations caused by the strong non-linearity of the underlying problem. The latter point can be explained by the derivations in the current paper, where the conventional DANN structure is shown to be optimal for Gaussian distribution of inputs (and also Gaussian signal processing on each node) that is hardly fulfilled in the considered CO$_2$ benchmark.

\section*{Summary and Conclusions}
The current paper analyzes the neural signal processing in deep artificial neural networks from the point of view of polynomial chaos (PCE) theory. The response on each node of a deep network has been seen from the PCE perspective, making use of orthonormal polynomial basis functions. The proposed generalization of the conventional structure of DANNs towards the DaPC NN decomposes the neural signals, employing an adaptive construction of data-driven multi-variate orthonormal bases on each node in the multi-layer structure. Moreover, by introducing PCE theory to represent the response of each node, it offers an additional opportunity to go beyond the linear weighted superposition of single \rw{univariate} neurons as in conventional DANN structures. In that sense, the introduced DaPC NN structure can be seen as a generalization of the conventional DANN with incorporation of aPC theory. The newly introduced DaPC NN assures orthonormal decomposition on each node and also offers an additional possibility to account for high-order neural effects. Doing so, the weights gain a clear meaning according to global sensitivity analysis and they reflect the partial contribution of each single neuron (linear \rw{univariate} terms) or simultaneous combination of neurons (non-linear \rw{multivariate} terms) to the total variance of the response on each node.

Concluding the analysis \rw{of} the current paper, we anticipate that avoiding the redundant representation and accounting for high-order neural effects could increase the performance of the neural network. Following the observation in the current paper, we stress that the high non-linearity of the underlying problems could limit the ability of the plain aPC with only one-layer polynomial representation. As opposed to that, the encoded flexibility of DANNs seems to suffer significantly from over-fitting, regardless of the regularization of weights. Overall, the DaPC NN shows \rw{the} ability to predict quantities of interest more consistently with lower variance in comparison to DANNs. Furthermore, we also observe that omitting the regularization in the loss function only slightly degrades the results of DaPC NN and aPC, whereas the conventional DANN is significantly more affected by it. 

Remarking that aspect, we accentuate that joining the fundamentals of homogeneous chaos theory with the deep representation of neural networks requires additional research, where various architectures accompanied by specific loss functions and training algorithms could be \rw{investigated.} \del{incorporated.} In particular, concepts employed in recurrent neural networks, neural ordinary differential equations, physical regularization or Bayesian regularization could be adopted directly to the DaPC NN structure. Moreover, the introduced DaPC NN structure opens a pathway to use analytical properties quantifying the importance of neural signal on each node \del{(e.g. sensitivity indices)}. \rw{In particular, implementing an explicit analytical form of data-driven orthonormal polynomial bases in equations (17)-(23) from \cite{OladNowak_RESS2012} could significantly accelerate the training process of DaPC NN. Additionally, since the weights of DaPC NN reflect the meaning according to global sensitivity analysis, they could be partially omitted \cite{burkner2022sparse} to offer a sparse DaPC NN representation. The latter is highly beneficial in mitigating issues related to overfitting.} Additionally, state-of-the-art findings in the PCE community can further be included, such as various sparse learning techniques, multi-element decomposition, Bayesian learning, etc. 

\rw{From a technical perspective, we would like to clarify that the primary goal of the current paper is not to achieve high computational efficiency in the training procedure. Instead, we offer a unique perspective on the DANN structure.}
\del{From a technical point of view, we would like to note that the current paper does not aim to achieve high computational efficiency of training procedure itself offering a first look at the DANN structure from a different angle.}
Indeed, the computational time of the current version of DaPC NN is significantly higher than that of DANN, since it requires adaptive computation of the orthonormal basis during the training process. The procedure could be accelerated using explicit analytic relations for a moderate degree of expansion from \cite{OladNowak_RESS2012}. Moreover, further speed up could be achieved when employing parallel and GPU-based computing. 

\section*{Acknowledgments}
The authors would like to thank the German Research Foundation (DFG) for the support of the project within the Cluster of Excellence "Data-Integrated Simulation Science" (EXC 2075) and Project Number 327154368 (SFB 1313) at the University of Stuttgart.

\end{document}